\documentclass{article} 
\usepackage{iclr2026_conference,times}

\usepackage{amsmath,amsfonts,bm}

\def\eqref#1{equation~\ref{#1}}

\def\1{\bm{1}}

\DeclareMathAlphabet{\mathsfit}{\encodingdefault}{\sfdefault}{m}{sl}
\SetMathAlphabet{\mathsfit}{bold}{\encodingdefault}{\sfdefault}{bx}{n}

\usepackage{hyperref}
\usepackage{url}
\usepackage{booktabs}       
\usepackage{amsfonts}       
\usepackage{amsmath}
\usepackage{nicefrac}       
\usepackage{microtype}      
\usepackage{xcolor}         
\usepackage{graphicx}
\usepackage{subcaption}
\usepackage{multirow}
\usepackage{enumitem}
\usepackage{wrapfig}

\title{SatDreamer360: Multiview-Consistent Generation of Ground-Level Scenes from Satellite Imagery}

\author{%
  Xianghui Ze\textsuperscript{1},  \quad Beiyi Zhu\textsuperscript{2}, \quad Zhenbo Song\textsuperscript{1}, \quad Jianfeng Lu\textsuperscript{1}, \quad Yujiao Shi\textsuperscript{2}\thanks{Corresponding author}\\
  \textsuperscript{1}Nanjing University of Science and Technology,  \textsuperscript{2}ShanghaiTech University\\
  \texttt{\{zexh, songzb, lujf\}@njust.edu.cn, \{v-zhuby,shiyj2\}@shanghaitech.edu.cn}
}

\iclrfinalcopy 
\begin{document}

\maketitle

\begin{abstract}
Generating multiview-consistent $360^\circ$ ground-level scenes from satellite imagery is a challenging task with broad applications in simulation, autonomous navigation, and digital twin cities. Existing approaches primarily focus on synthesizing individual ground-view panoramas, often relying on auxiliary inputs like height maps or handcrafted projections, and struggle to produce multiview consistent sequences. In this paper, we propose SatDreamer360, a framework that generates geometrically consistent multi-view ground-level panoramas from a single satellite image, given a predefined pose trajectory. To address the large viewpoint discrepancy between ground and satellite images, we adopt a triplane representation to encode scene features and design a ray-based pixel attention mechanism that retrieves view-specific features from the triplane. To maintain multi-frame consistency, we introduce a panoramic epipolar-constrained attention module that aligns features across frames based on known relative poses. 
To support the evaluation, we introduce {VIGOR++}, a large-scale dataset for generating multi-view ground panoramas from a satellite image, by augmenting the original VIGOR dataset with more ground-view images and their pose annotations. Experiments show that SatDreamer360 outperforms existing methods in both satellite-to-ground alignment and multiview consistency.
\end{abstract}

\section{Introduction}

Generating ground-level scenes from satellite imagery has attracted significant attention due to the broad coverage and low acquisition cost of satellite images. This task shows promising applications in autonomous driving~(\cite{villalonga2021leveraging, lu2024generative}), 3D reconstruction~(\cite{liu20243dgs, yan2024streetcrafter})  and data augmentation~(\cite{yang2023bevcontrol,gao2023magicdrive}) for downstream tasks. Many existing works (\cite{li2024crossviewdiff, lin2024geometry, xu2024geospecific, ze2025controllable}) focus on generating individual ground images from satellite views, leaving the continuity of multi-ground views largely unaddressed. In this paper, we aim to synthesize multiple ground-view images from a single satellite image, controlled by a predefined trajectory. This introduces new challenges in maintaining both geometric consistency with the top-down satellite image and multiview coherence across the sequence of generated frames.

Early approaches~(\cite{isola2017image, regmi2018cross, shi2022geometry, lu2020geometry, qian2023sat2density}) formulate cross-view synthesis as a one-to-one mapping problem, often implemented with Conditional Generative Adversarial Networks (cGANs). These methods focus on aligning representations at pixel or perceptual level. However, the extreme viewpoint disparity between top-down satellite views and street-level images leads to limited field-of-view overlap. Satellite images inherently miss key elements such as building facades, tree trunks, and other occluded details, making the ground view generation task highly under-constrained and naturally one-to-many.

Recent advances leverage latent diffusion models (LDMs)~(\cite{rombach2022high}) to better handle this uncertainty~(\cite{li2024crossviewdiff, lin2024geometry, deng2024streetscapes, xu2024geospecific, ze2025controllable}). These methods introduce probabilistic modeling to produce diverse and high-fidelity ground images. However, they often rely on approximate projections~(\cite{lin2024geometry, ze2025controllable}) or auxiliary data such as height maps~(\cite{li2024crossviewdiff, deng2024streetscapes, xu2024geospecific}), which can be difficult to obtain at scale. Moreover, while effective for single-view generation, these models fall short in producing multiview consistent sequences, which are critical for applications like simulation, planning, or digital twin city modeling. A recent effort~(\cite{xu2025satellite}) attempts to generate continuous ground-view videos by leveraging multi-angle satellite imagery in a two-stage pipeline: the first stage generates a base frame, followed by autoregressive generation of future frames. While this improves continuity, the reliance on multi-view satellite input and complex coordination for different generation stages reduces practical applicability.

\begin{figure}[t]
    \centering
    \setlength{\abovecaptionskip}{0pt}
    \setlength{\belowcaptionskip}{0pt}
    \includegraphics[width=1\textwidth]{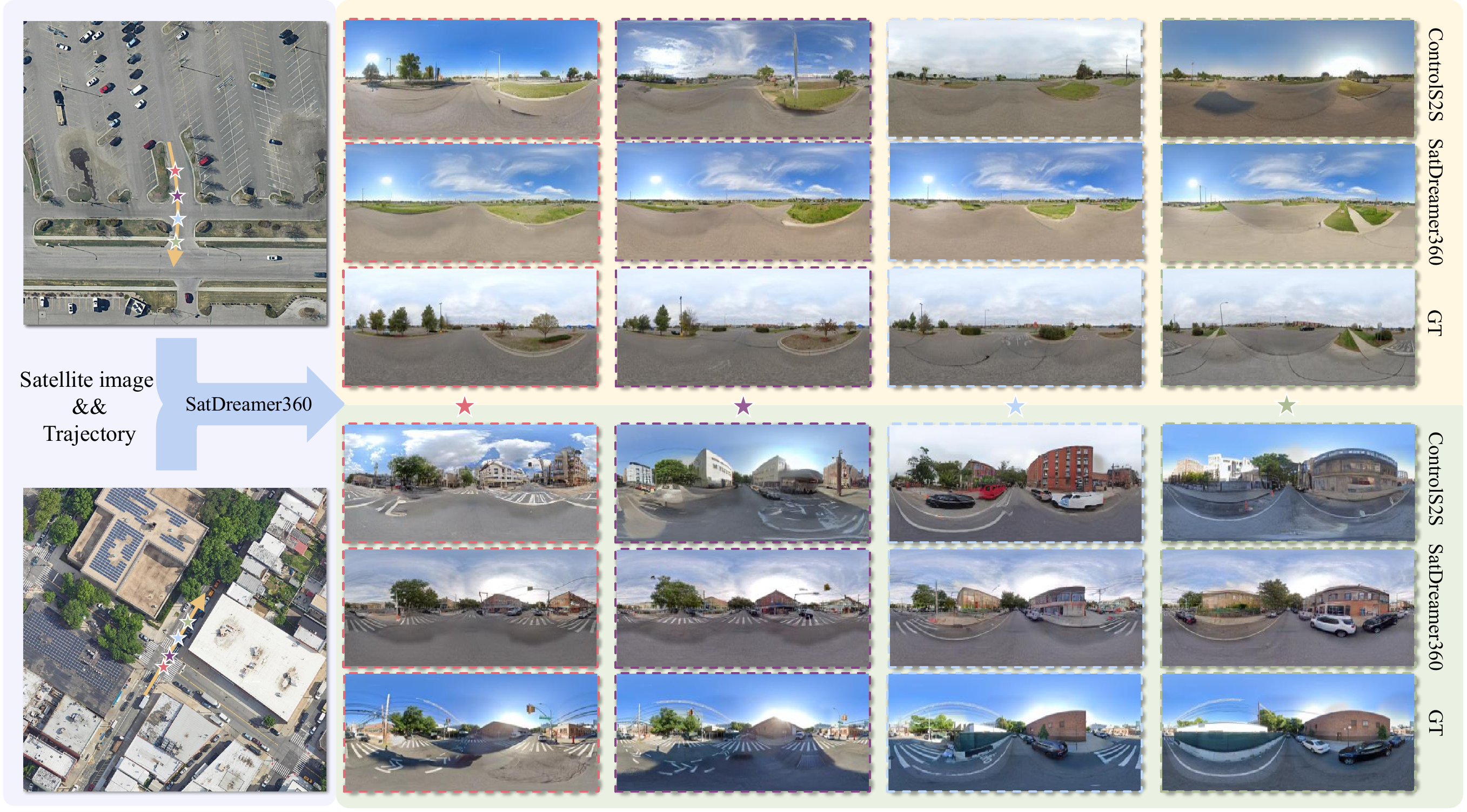}
    \caption{\small
    Given a satellite image and a sequence of query poses (colored stars), our goal is to synthesize coherent panoramic views along the trajectory. The proposed SatDreamer360 generates more realistic and geometrically consistent ground-level scenes compared to state-of-the-art methods, faithfully capturing spatial layouts and structural continuity across diverse environments.
    \label{Figure:head}}
    \vspace{-2em}
\end{figure}

In this paper, we present SatDreamer360, a unified framework that generates continuous and coherent ground-view sequences from a single satellite image and a target trajectory, as shown in Figure~\ref {Figure:head}. The key idea is to embed explicit cross-view geometric reasoning between satellite and ground views, as well as across ground frames into the latent diffusion process. 

We adopt a compact triplane representation~(\cite{huang2023tri,bhattarai2024triplanenet,shue20233d}) to encode scene geometry directly from the satellite image, avoiding the need of height maps~(\cite{deng2024streetscapes, xu2025satellite}) or handcrafted projections~(\cite{ze2025controllable}). We further design a ray-based pixel attention mechanism that retrieves view-dependent features from the triplane and integrates them into conditional diffusion, enabling geometry-aware and controllable generation.

To enhance mutiview consistency, we draw inspiration from the use of epipolar constraints in pinhole cameras~(\cite{tobin2019geometry,he2020epipolar,huang2022h,huang2024epidiff}) and extend the idea to panoramic images with equirectangular projections. We design an epipolar-constrained attention module for panoramic images, which aligns features across frames by leveraging the known relative camera poses.

Finally, to support large-scale evaluation, we construct VIGOR++, an extension of the VIGOR dataset with ground-truth trajectories and continuous ground-view sequences, providing a new benchmark for cross-view generation. To summarize, our contributions are as follows:
\begin{itemize}[noitemsep,topsep=1pt]
\item A unified framework, SatDreamer360, for generating continuous and geometrically consistent ground-view sequences from a single satellite image and a target trajectory.

\item A ray-guided cross-view feature condition mechanism that encodes the 3D scene with a triplane representation and aggregates spatial cues pixel by pixel via ray-based attention, enabling geometry-aware and controllable diffusion-based ground-view generation.

\item An interframe attention module that uses panoramic epipolar constraints via equirectangular projections to align features across frames, enhancing multiview consistency.

\item A new VIGOR++ dataset, which extends VIGOR with continuous sequences and trajectory annotations, providing a benchmark for cross-view sequence synthesis.




\end{itemize}

\section{Related work}
\textbf{Cross-view ground scene generation} aims to reconstruct ground scenes from other perspectives, such as aerial~(\cite{xu2023grid, gao2024skyeyes}) or landmark images~(\cite{yang2023bevcontrol,li2024drivingdiffusion, gao2023magicdrive, swerdlow2024street}). Given the wide availability of satellite imagery, related research~(\cite{shi2022geometry, qian2023sat2density, xu2024geospecific, ze2025controllable,li2021sat2vid, li2024sat2scene, xu2025satellite}) focuses on satellite-to-ground generation. Previous works~(\cite{isola2017image, regmi2018cross}) implicitly convert satellite image features into ground map representations, often causing geometric distortions. 
Later methods introduced approximate projections~(\cite{lu2020geometry, shi2022geometry, lin2024geometry, ze2025controllable}), height maps~(\cite{mari2022sat, lu2020geometry, deng2024streetscapes,xu2024geospecific, li2024crossviewdiff, xu2025satellite}), or estimating density maps~(\cite{qian2023sat2density}) as priors. However, their accuracy is constrained by errors in the estimation priors, and ground-image-based methods often over-rely on these inaccurate projections while neglecting the broader contextual information available from satellite imagery.

\textbf{Multiview consistent image generation} aims to generate multiview continuous frames from given prompts. Early Gan-based approach~(\cite{vondrick2016generating, rematas2022urban}) has been surpassed by methods utilizing the diffusion architecture~(\cite{blattmann2023align, blattmann2023stable}), where Video Diffusion Models (VDM)~(\cite{blattmann2023align,singer2022make,wu2023tune}) introduce spatiotemporal modules into U-Net to generate coherent sequences, though with high computational cost. Recent works~(\cite{tseng2023consistent,huang2024epidiff}) incorporate epipolar constraints to enforce multiview consistency, but are mostly limited to pinhole cameras, with little exploration of panoramic settings. More broadly, methods such as~(\cite{liu2023zero, kong2024eschernet, voleti2024sv3d, bourigault2024mvdiff}) generate multi-view images from single-object inputs, yet they focus on object-level generation and cannot address scene-level continuity.
\section{Method \label{sec:Method}}

Given a satellite image $S$ and a set of 4-DoF ground camera poses $\{p^i = [t^i, \psi^i]\}$, where $t^i$ denotes spatial location and $\psi^i$ the yaw angle, our goal is to synthesize a sequence of ground panoramic images ${G^i}$ that are spatially aligned with the satellite view and consistent across multiple views.

\begin{figure}[t]
    \centering
    \includegraphics[width=1\textwidth]{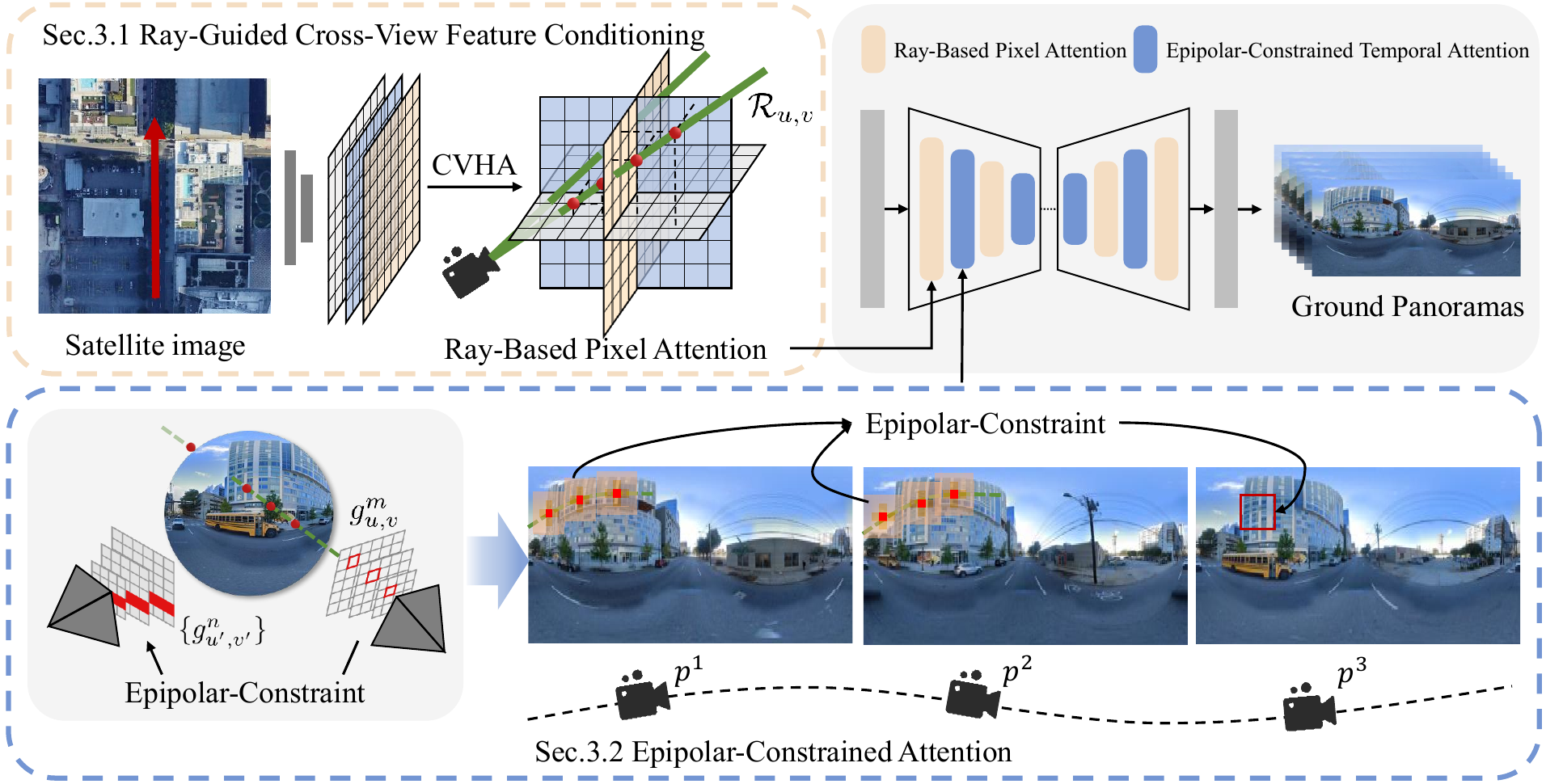}
    \caption{\small 
    Overview of the proposed SatDreamer360 framework. Given a single satellite image and a target trajectory, our model synthesizes continuous ground-level panoramas along the path. A Ray-Based Pixel Attention mechanism retrieves view-specific features through cross-view geometric reasoning, guided by a tri-plane representation of the scene. An Epipolar-Constrained Attention module aligns features across frames using relative camera poses. 
    }
    \label{Figure:framework}
    \vspace{-1.7em}
\end{figure}

To obtain the optimal solution for ground image inference, as illustrated in Figure~\ref{Figure:framework}, we develop SatDreamer360 based on a latent diffusion model~(\cite{song2020score,blattmann2023align}) to synthesize ground-level views conditioned on the satellite image and camera pose. It generates ground images by iteratively denoising a random Gaussian noise for $T$ steps, learning to predict the Gaussian noise $\epsilon$ injected at each step $t$:
\begin{equation}
    \mathcal{L} = \mathbb{E}_{z_0, c, \epsilon, t} \left[\| \epsilon - \epsilon_{\theta}(\sqrt{\bar{\alpha}_t}z_0 + \sqrt{1 - \bar{\alpha}_t}\epsilon, t, c) \|^2 \right].
\end{equation}
Here, $\epsilon_\theta$ is the denoising network using U-Net, $c$ is the conditioning input—comprising the satellite image $S$ and pose $p^i$, $\bar{\alpha}_t$ is the variance schedule, and $\epsilon$ is drawn from a standard Gaussian distribution. Ground images $G$ are encoded using a VQ-VAE~(\cite{esser2021taming}) encoder $\mathcal{E}(G)$ to obtain latent codes $z$. For clarity, we refer to ground representations in latent space also as $G$ in what follows.

\subsection{Ray-Guided Cross-View Feature Conditioning}

\textbf{Spatial Representation via Triplanes.}
To represent the 3D scene covered by the satellite image, we adopt a tri-plane structure~(\cite{chan2022efficient,huang2023tri}), a lightweight and expressive alternative, instead of the information-sparse BEV representation~(\cite{li2024bevformer}) or the computationally intensive voxel representation~(\cite{li2022unifying}). Three orthogonal planes $(XY, XZ, YZ)$ are defined in the tri-plane representation, with the $XY$ plane parallel to the ground. 

Given a point in 3D space, its feature $F_{xyz}$ is obtained by aggregating the features from its projections onto the three planes:
\begin{equation}
    F_{xyz} = F_{xy} \oplus F_{xz} \oplus F_{yz},
    \label{eq:triplane_3d_feat}
\end{equation}
where $F_{xy}$, $F_{xz}$, and $F_{yz}$ denote the interpolated features from the corresponding 2D planes, and $\oplus$ denotes their element-wise summation to obtain the final 3D feature.

To construct the triplane representation, we initialize the planes by extracting features from the satellite image using a ResNet~(\cite{he2016deep}), which naturally aligns with the top-down $XY$ plane. To enrich spatial reasoning across all three orthogonal planes, we apply Cross-view Hybrid Attention (CVHA~\cite{li2024bevformer}), enabling interactions among the $XY$, $XZ$, and $YZ$ planes. Each plane aggregates projections from the other two, enriching its features with complementary spatial context. For instance, the updated features on the $XY$ plane are computed as:
\begin{equation}
    F_{xy}^{\text{top}} = \mathrm{CVHA}\left(F_{xy}^{\text{top}}, \mathrm{Ref}_{xy}^{3D}\right), \quad \mathrm{Ref}_{xy}^{3D} = F_{xy}^{\text{top}} \cup \{F_{yz_i}^{\text{side}}\} \cup \{F_{xz_i}^{\text{front}}\}.
\end{equation}
Here, $F_{xy}^{\text{top}}$ denotes the point feature on the $XY$ plane. 
The reference set $\mathrm{Ref}_{xy}^{3D}$ contains local neighbors sampled along the $Z$-axis from the orthogonal $XZ$ and $YZ$ planes, denoted as $\{F_{xz_i}^{front}\}$ and $\{F_{yz_i}^{side}\}$. This cross-plane aggregation enables each point on the triplane to incorporate multiview cues, thereby enhancing 3D spatial consistency.
Moreover, in sequential settings, previously synthesized ground-view images can be projected back and integrated into the triplane to refine its representation. With CVHA, this incremental update yields a more expressive and temporally coherent scene model. Further architectural and implementation details are provided in the Appendix~\ref{App:triplane_updare}.


\textbf{Ray-Based Pixel Attention.}
Conventional cross-attention mechanisms~(\cite{rombach2022high}) typically align global prompts with image-level semantics but often fail to respect underlying 3D scene geometry. This limits their ability to establish accurate cross-view correspondences, particularly in view synthesis tasks. To address this, we propose a Ray-Based Pixel Attention module that incorporates geometric priors by explicitly conditioning attention on camera rays.

Specifically, as illustrated in Figure~\ref{Figure:framework} (top middle) and Appendix~\ref{App:Projection}, each pixel $g_{u,v}$ at location $(u, v)$ in the panoramic ground-view image $G \in \mathbb{R}^{H \times W \times C}$ corresponds to a unique 3D ray $\mathcal{R}_{u,v}$, parameterized by yaw $\psi$ and pitch $\theta$ angles:
\begin{equation}
    \psi_{u,v} = (u - \frac{W}{2})/W \times 2\pi , \quad
    \theta_{u,v} = (\frac{H}{2} - v)/H \times \pi .\label{eq:cam_project}
\end{equation}
These angular parameters define the direction of the ray $\mathcal{R}_{u,v}$ in the camera coordinate system. The ray originates from the ground-view camera position, and its direction is uniquely defined by $(\psi_{u,v}, \theta_{u,v})$. To encode spatial cues along each ray, we sample $K$ points at evenly spaced depths $\{r_k\}_{k=1}^K$, and project them into the spatial coordinate system using the camera pose, yielding 3D positions $\mathbf{x}_{u,v,k}$. Features at these 3D positions are then extracted from the tri-plane representation using deformable attention:
\begin{equation}
    F_{g_{(u,v)}} = \sum_{j=1}^{J} W_j \sum_{k=1}^{K} A_{k,j} \cdot F_{\left( \mathbf{x}_{u,v,k} + \Delta \mathbf{x}_{k,j} \right)},
    \label{eq:Ray_att}
\end{equation}
where $J$ is the number of attention heads, $W_j$ is a learnable weight for head $j$, $\Delta \mathbf{x}_{k,j}$ is an offset along the ray around the sampled points, initialized to zero, and $A_{k,j}$ denotes the attention weight for these sampled points, normalized such that $\sum_{k=1}^{K} A_{k,j} = 1$ for each head. Both offsets and attention weights are dynamically refined across iterations, guided by the evolving ground latent feature map. $F_{\left( \mathbf{x}_{u,v,k} + \Delta \mathbf{x}_{k,j} \right)}$ denotes features extracted from the triplane at the adjusted 3D positions using Eq.~\ref{eq:triplane_3d_feat}. The aggregated feature $F_{g_{(u,v)}}$ at pixel $(u,v)$ guides the U-Net in integrating satellite information, effectively aggregating spatial cues pixel by pixel.

\begin{figure}[t]
    \centering
    \includegraphics[width=1\textwidth]{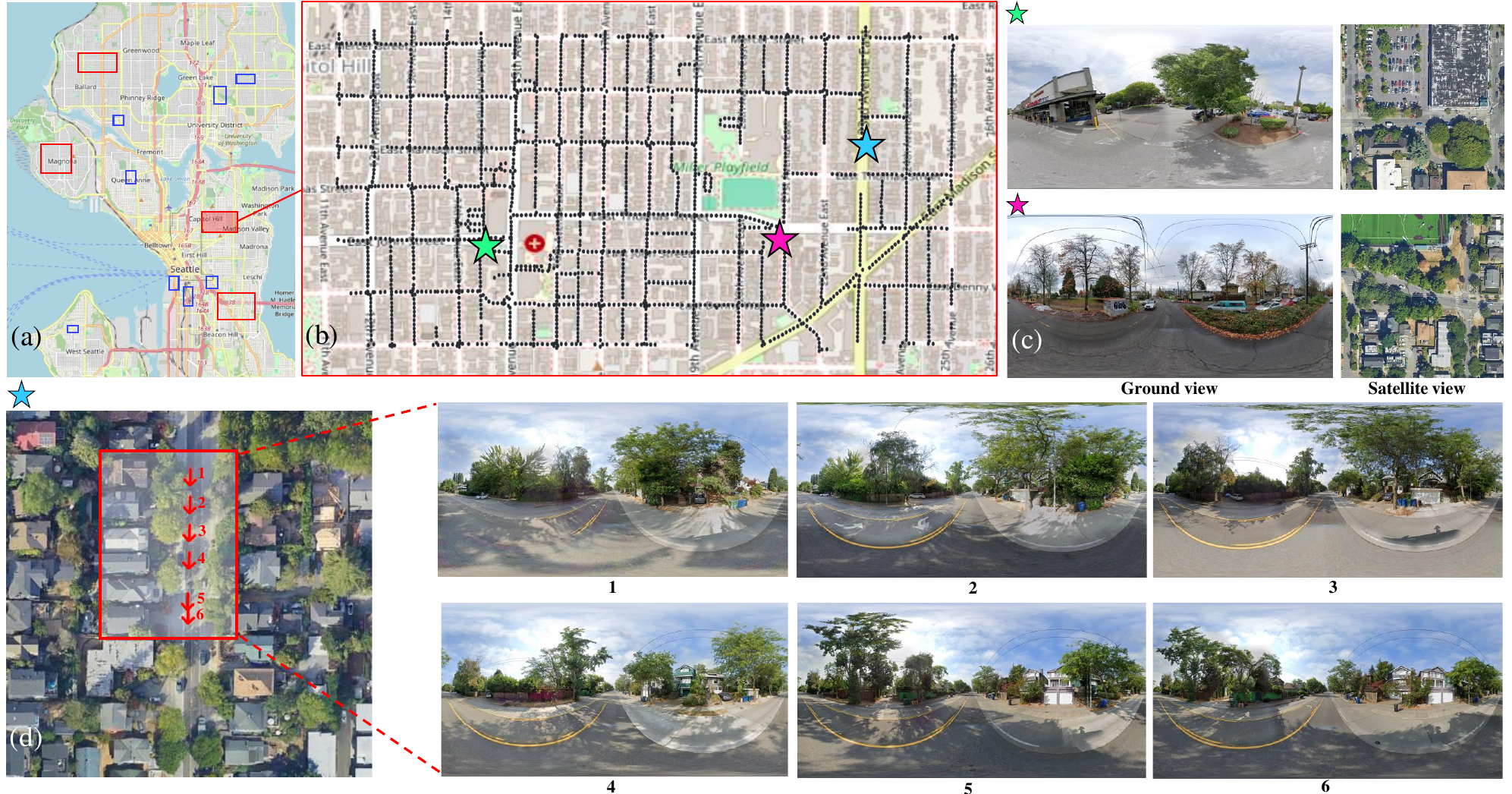}
    \caption{\small Overview of the VIGOR++ dataset. (a) The map of Seattle, USA, serves as an example of the ten cities in the dataset. The red boxes and blue boxes represent the districts for the training set and test set, respectively. (b) shows a road map. Dots and stars along the road represent locations of ground images and satellite images. Two of them, marked with the red star and green star, are shown in (c). (d) shows the continuous ground sequence within one satellite image.}
    \label{fig:dataset}
    \vspace{-1.7em}
\end{figure}

\subsection{Epipolar-Constrained Attention}
To maintain mutiview consistency across consecutive frames in a lightweight and efficient manner, we draw inspiration from epipolar-constrained attention in pinhole images~(\cite{tseng2023consistent}) and extend it to outdoor panoramic imagery. Specifically, we introduce an attention mechanism tailored for equirectangular projections. For two frames of ground panoramas, $G^m$ and $G^n$, a pixel $g^m_{u,v}$ on $G^m$ corresponds to a set of candidate pixels on $G^n$, enforced by the underlying geometric constraint (proof in Appendix~\ref{App:epipolar_proof}):
\begin{equation}
    \left(P^{-1}(g^n_{u',v'})\right)^\top \hat{t}_{mn} R_{mn} \left(P^{-1}(g^m_{u,v})\right) = 0. \label{eq:epipolar}
\end{equation}
Here, the point set $\{g^n_{u',v'}\}$ on $G^n$ denotes candidate matches that satisfy the constraint relationships. The terms $R_{mn}$ and $t_{mn}$ denote the relative rotation and translation between frames $m$ and $n$, and $\hat{t}_{mn}$ is the skew-symmetric matrix of $t_{mn}$. The mapping  $P$ is the equirectangular camera projection defined in Eq.~\ref{eq:cam_project}. Therefore, when establishing mutiview consistency, we do not need to perform pixel-wise correspondence for the entire image as in previous work~(\cite{wu2023tune,xu2025satellite}). Instead, we restrict attention to points that satisfy the epipolar constraint, significantly reducing redundancy while preserving geometric fidelity:
\begin{equation}
    F_{g^m_{u,v}} = \mathrm{softmax}\left(\frac{QK^\top}{\sqrt{d}}\right) V, Q = W^QF_{g^m_{u,v}}, K = W^KF_{\{g^n_{u',v'}\}}, V = W^VF_{\{g^n_{u',v'}\}},
\end{equation}
where $W^Q$, $W^K$, and $W^V$ are learnable matrices for query, key, and value. This epipolar-constrained attention is applied at multiple U-Net levels to fuse coarse and fine-grained features. By restricting attention to points that satisfy the epipolar constraint, the computational complexity is reduced from $O(NHW \times NHW)$ to $O(NHW \times NM)$, where $N$ is the number of frames in sequences, $H$ and $W$ denote the height and width of the feature map, and $M$ is the number of sampled points satisfying epipolar constraints with $M \ll HW$. To further improve efficiency, we adopt a sparse strategy, querying only the two preceding frames to balance coherence and cost.

\subsection{VIGOR++: Extending VIGOR for Satellite-to-Ground Video Generation}
Existing cross-view datasets lack continuous panoramic sequences. To address this, we construct VIGOR++, an extension of the VIGOR dataset~(\cite{zhu2021vigor}) tailored for large-scale, consistent cross-view generation, enabling the dataset to be more widely used in 3D scene reconstruction, cross-view video localization tasks, as shown in Figure~\ref{fig:dataset}.
To broaden the coverage of satellite maps for the task of large-scale scene generation, we expand the wide-area satellite map dataset by increasing it from the original $70\,\mathrm{m} \times 70\,\mathrm{m}$ to $200\,\mathrm{m} \times 200\,\mathrm{m}$ from Google Maps~(\cite{googlemaps_maps}). Subsequently, we include additional cities. Apart from the initial cities of Chicago, New York, San Francisco, and Seattle, we integrate datasets for six additional regions: Atlanta, Bismarck, Kansas, Nashville, Orlando, and Phoenix. This augmentation enriches the variety of urban representations within the dataset.
To obtain continuous ground sequences, we extract all available Google Street View~(\cite{googlemaps_streetview}) images within the satellite region. Subsequently, we employed a semi-automatic approach to organize sampling paths for each satellite image. By leveraging sky color histograms and image embedding similarities, we constructed a connectivity graph and executed path extraction based on depth-first search to identify potential routes. Subsequent manual refinement ensured multiview coherence. Our efforts yielded more than 90,000 novel cross-view satellite and ground video pairs. Of these, 84,055 pairs are designated for training, while 7,443 are allocated for testing. To evaluate the model's generalization capabilities, the testing set is collected from locations entirely distinct from the training data.

\section{Experiments \label{sec:Experiments}}
\textbf{Experimental Setup.}  
We use $256 \times 256$ satellite images and the 4-DoF camera poses of ground-view images as input, aiming to generate continuous ground-view sequences at a resolution of $128 \times 512$ for fair comparison with prior work. Our model is finetuned based on the pre-trained Stable Diffusion 1.5 model~(\cite{rombach2022high}). We first perform 300 epochs of finetuning on a single-image generation task, followed by an additional 300 epochs on continuous sequence data dataset to learn temporal consistency. During inference, we adopt DDIM sampling with 50 steps for efficient generation. All experiments are conducted using four NVIDIA L40 GPUs.

\textbf{Datasets.}  
For the single ground-view image generation task, we use the CVUSA~(\cite{zhai2017predicting}) dataset, which primarily focuses on rural areas, and VIGOR~(\cite{zhu2021vigor,lentsch2022slicematch}) dataset, which covers four major cities, following the same protocol as prior works~(\cite{shi2022geometry, qian2023sat2density, ze2025controllable}). These cross-view datasets provide one-to-one correspondences between panoramic ground images and satellite images. CVUSA contains 35,532 pairs for training and 8,884 pairs for testing. VIGOR contains 52,609 pairs for training and 52,605 for testing. For the continuous scene generation task, experiments are conducted using our proposed VIGOR++ dataset. 


\textbf{Evaluation Metrics.}  
We evaluate the authenticity and multiview consistency of generated images. For authenticity, we compare results with ground truth (GT) using pixel-level metrics (SSIM, PSNR, SD) and perceptual metrics based on pretrained networks ($P_{\text{alex}}$~(\cite{krizhevsky2012imagenet}),  $P_{\text{squeeze}}$~(\cite{iandola2016squeezenet}), and FID~(\cite{heusel2017gans})). 
To account for real-world variations in weather and season, which can cause color shifts, we focus on structural and semantic similarity using DINO~(\cite{caron2021emerging}), Segment Anything~(\cite{kirillov2023segment}), and DepthAnything~(\cite{yang2024depth}) for depth consistency. In addition, we employ LRCE~(\cite{shen2022panoformer}) to measure the continuity of panoramic images along the left and right boundaries. Multiview consistency across frames is assessed with FVD~(\cite{unterthiner2018towards}) and CLIPSIM~(\cite{wu2021godiva}) to measure sequence coherence and stability. Additionally, to further evaluate multiview geometric consistency, we perform 3D reconstruction using the generated views and compare the results with reconstructions synthesized from GT views. Specifically, we employ Pi3~(\cite{wang2025pi}), a feed-forward 3D reconstruction network, to reconstruct the scene from image sequences, and then render images from the reconstructed 3D volumes using the same camera views. The rendered views are compared with the ground truth using standard image similarity metrics, including PSNR and SSIM, denoted as 3DPS and 3DSS.


\subsection{Comparison with Prior Work on Satellite-to-Ground Sequence Generation}
Generating continuous and coherent ground-level Sequences from a single satellite image is highly challenging due to the extreme viewpoint gap and inherent spatial ambiguity. We compare our method against three representative baselines: Sat2Density~(\cite{qian2023sat2density}) and ControlS2S~(\cite{ze2025controllable}), both designed for cross-view image generation, and EscherNet~(\cite{kong2024eschernet}), a recent diffusion-based model for general multiview synthesis.
Since neither the code nor data of \cite{deng2024streetscapes,xu2025satellite} is released, comparisons with them cannot be conducted and thus are not included. 
Moreover, their settings rely on multiple satellite images from different viewpoints or real ground-truth height maps, while ours requires only a single overhead satellite image as input.

\begin{figure}[t]
    \centering
    \setlength{\abovecaptionskip}{0pt}
    \setlength{\belowcaptionskip}{0pt}
    \includegraphics[width=1\textwidth]{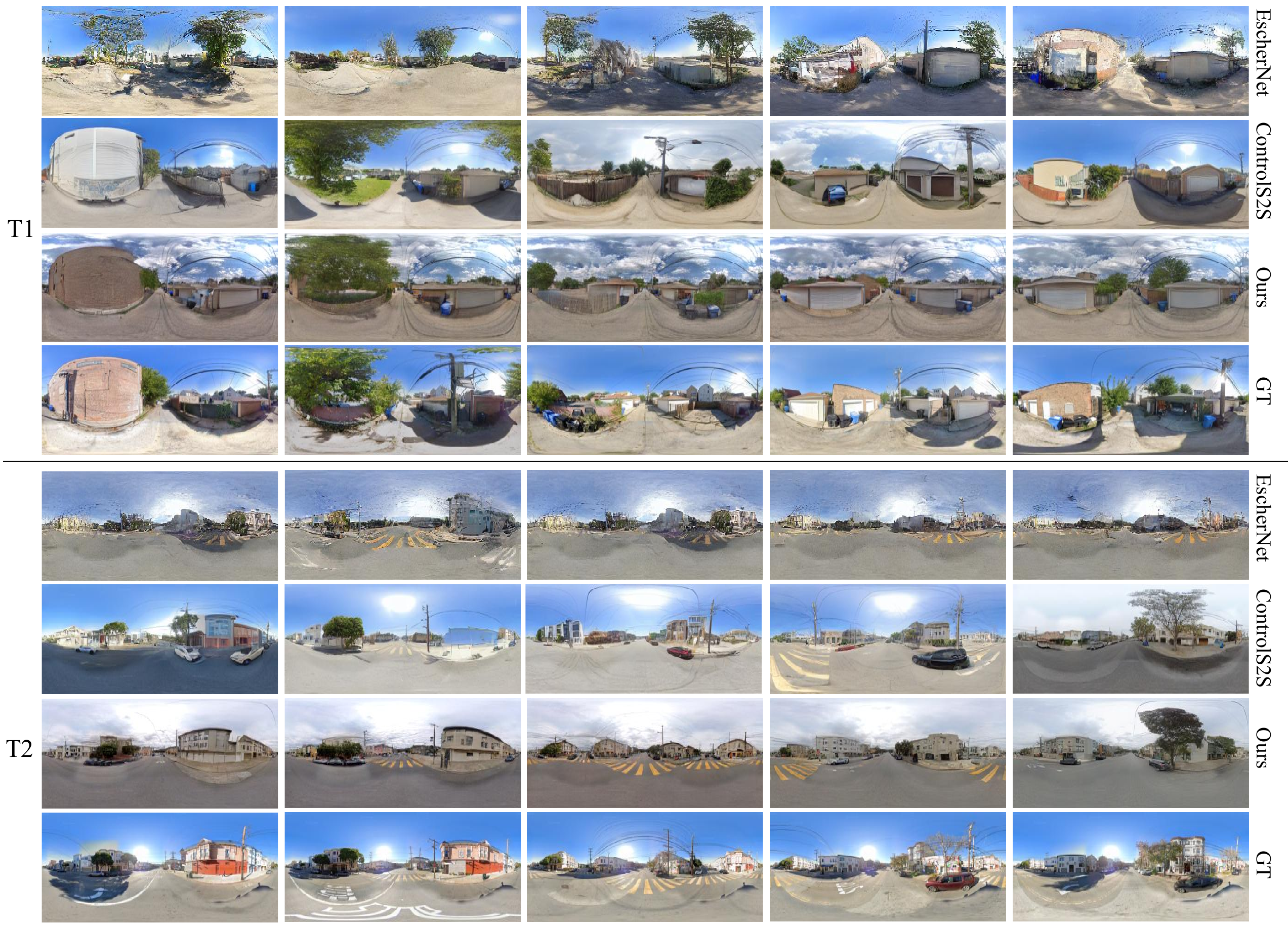}
    \caption{\small  
        Qualitative comparison of ground-level image sequences along trajectories T1 and T2, shown from left to right. The corresponding satellite images and trajectories are provided in Figure~\ref {fig:sat_traject}.
        Our method produces more realistic textures and preserves structural and spatial continuity across frames, demonstrating stronger multiview coherence and environmental fidelity across diverse scenes.
        \label{fig:sequence_vis}}
    \vspace{-1em}
  \end{figure}
  \begin{table}[t]
  \centering
  \fontsize{8.5}{10}\selectfont
  \setlength{\abovecaptionskip}{0pt}
  \setlength{\belowcaptionskip}{0pt}
  \caption{\small Quantitative comparison with existing algorithms on VIGOR++ dataset.\label{tab:VIGOR_seq_comp_sota}}
  \setlength{\tabcolsep}{0.8mm}{
  \begin{tabular}{c|cc|cc|cc|cccc|c}
\midrule
\multirow{2}{*}{Method} & \multicolumn{2}{c|}{Perceptual level}    & \multicolumn{2}{c|}{Semantic level}   & \multicolumn{2}{c|}{Pixel level} & \multicolumn{4}{c|}{Multiview level}                                     & \multirow{2}{*}{$\downarrow$Depth} \\
                        & $\downarrow$$P_{alex}$ & $\downarrow$FID & $\downarrow$DINO & $\downarrow$SegAny & $\uparrow$SSIM  & $\uparrow$PSNR & $\downarrow$FVD & $\downarrow$CLIPSIM & $\uparrow$3DPS   & $\uparrow$3DSS    &                                    \\ \midrule
Sat2Den                 & 0.4584                 & 133.6           & 4.437            & 0.3729             & 0.3892          & 12.06          & 11.70           & 8.405               & 11.16          & 0.4868          & 7.671                              \\
EscherNet               & 0.5581                 & 84.21           & 4.942            & 0.3845             & 0.2587          & 11.23          & 7.282           & 8.250               & 11.09          & 0.3591          & 10.50                              \\
ControlS2S              & 0.4433                 & 29.48           & 4.567            & 0.3753             & 0.3718          & 11.84          & 4.871           & 10.81               & 11.86          & 0.4647          & 6.651                              \\
Ours                    & \textbf{0.3955}        & \textbf{27.41}  & \textbf{4.156}   & \textbf{0.3563}    & \textbf{0.3964} & \textbf{12.75} & \textbf{2.101}  & \textbf{6.820}      & \textbf{13.31} & \textbf{0.5196} & \textbf{5.623}                     \\ \midrule
\end{tabular}
        }
      \vspace{-2em}
  \end{table}

Among these baselines, Sat2Density represents a Nerf-based method for cross-view generation.
ControlS2S is a recent diffusion-based method that synthesizes ground-level images conditioned on a single satellite image.
EscherNet is a state-of-the-art diffusion framework for multiview image generation. We adapt it to our setting by treating the satellite image as the source view and synthesizing the ground-view frames as target views.
For fair comparison, we retrain all methods on our proposed VIGOR++ dataset.

As shown in Table~\ref{tab:VIGOR_seq_comp_sota}, EscherNet performs the worst across perceptual, semantic, pixel-wise, and depth-consistency metrics, mainly because it lacks an explicit mechanism to bridge the large domain gap between satellite and ground views. However, it achieves better multiview consistency (measured by FVD and CLIPSIM) than Sat2Density and ControlS2S, owing to its built-in multi-view coherence modeling.
In contrast, SatDreamer360 explicitly addresses both cross-view appearance disparity and the challenge of multiview continuity. As a result, it achieves the best overall performance across all dimensions, combining high image fidelity with smooth and consistent video generation.
Qualitative results in Figure~\ref{fig:sequence_vis} further support these findings. EscherNet, which relies on implicit scene encoding, struggles to produce realistic ground-level images. ControlS2S, as illustrated in Figure~\ref{Figure:head}, lacks effective mechanisms for multiview consistency, leading to spatial discontinuities across frames. In comparison, SatDreamer360 preserves the underlying scene layout and produces ground-view sequences that are both spatially coherent and temporally smooth.
\begin{figure*}[t]
    \centering
    \setlength{\abovecaptionskip}{0pt}
    \setlength{\belowcaptionskip}{0pt}
    \begin{subfigure}{0.088\linewidth}
        \centering
        \includegraphics[width=\linewidth]{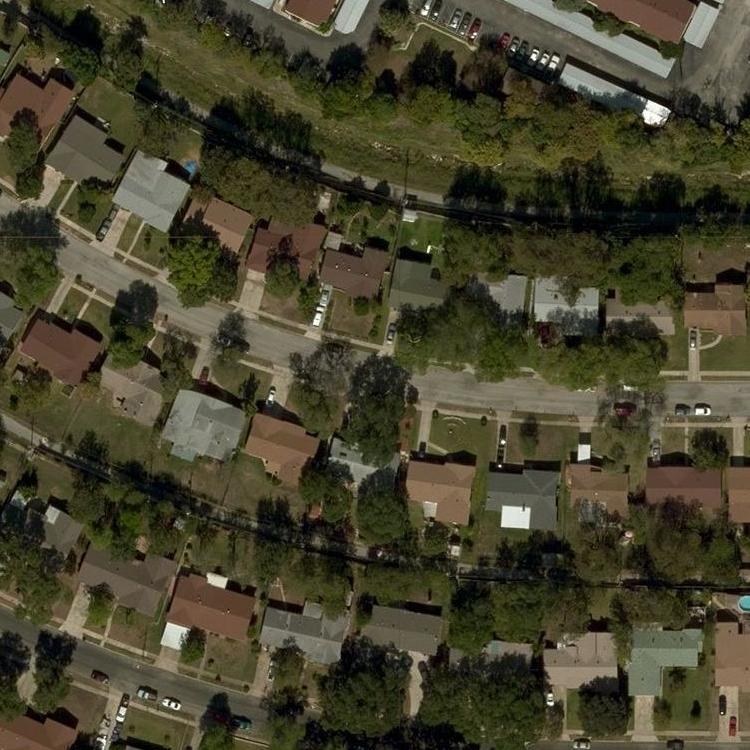}
        \includegraphics[width=\linewidth]{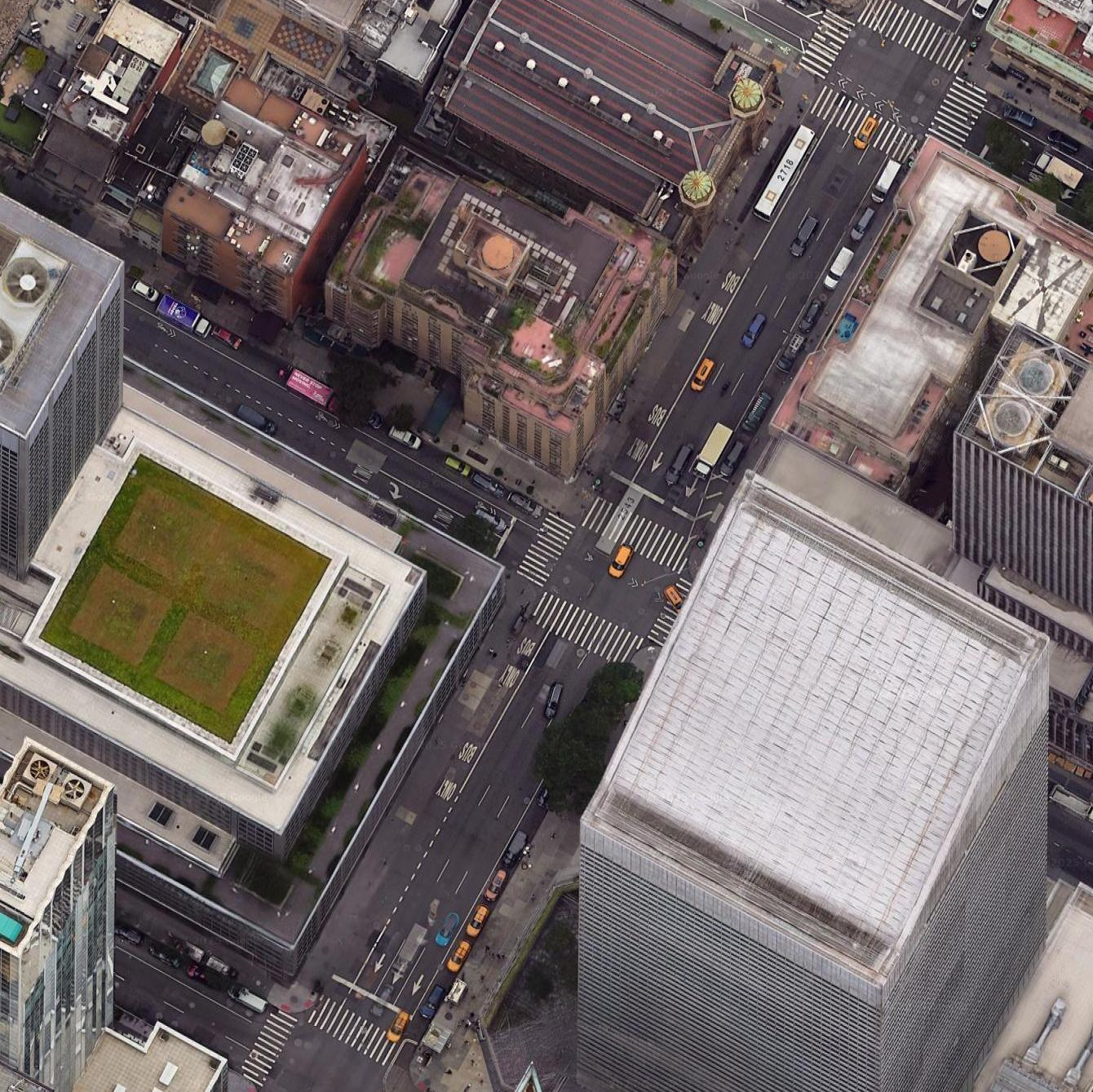} 
        \includegraphics[width=\linewidth]{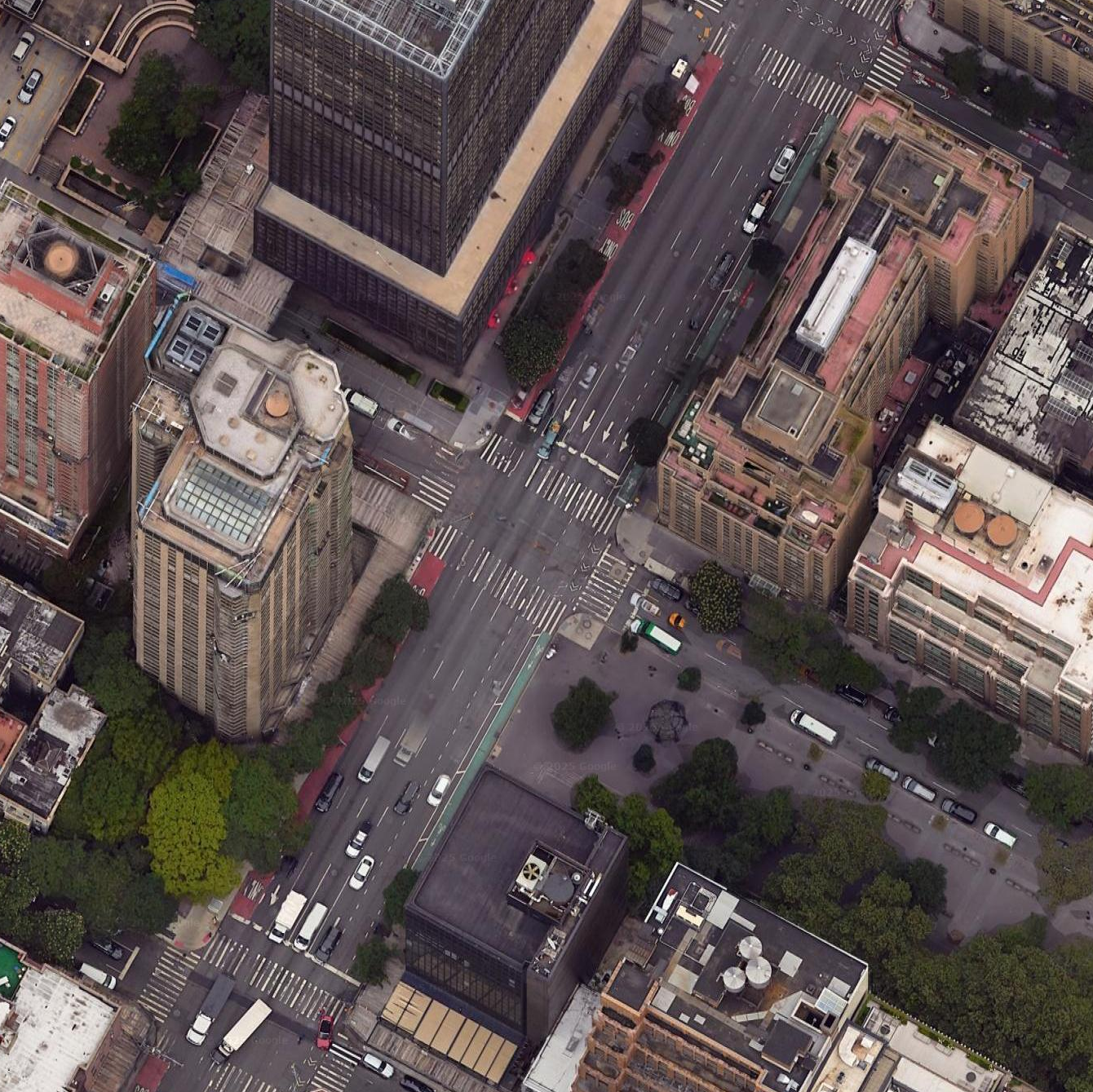} 
        \includegraphics[width=\linewidth]{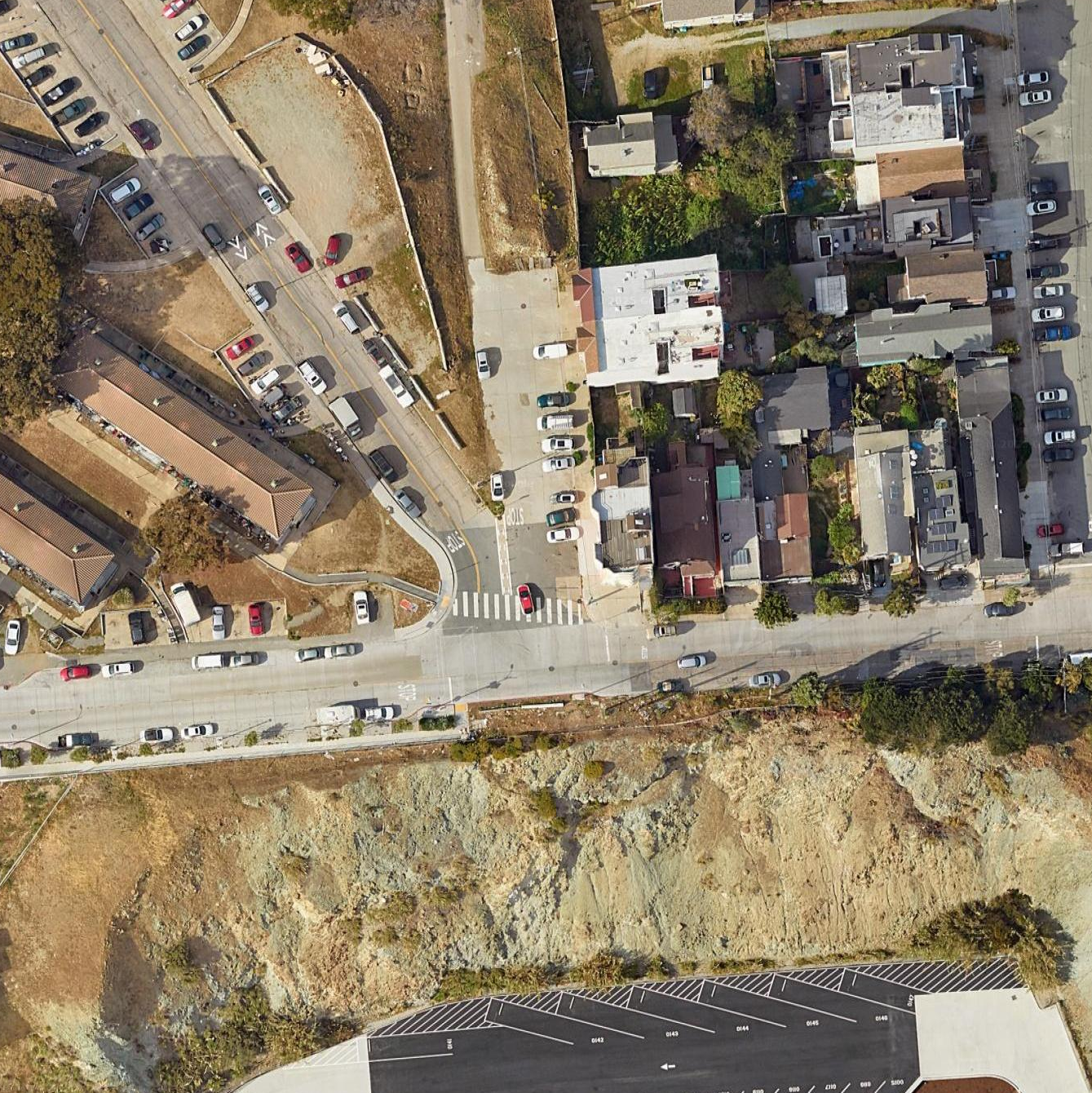} 
        \caption{\small Sat}
    \end{subfigure}
    \begin{subfigure}{0.22\linewidth}
        \includegraphics[width=\linewidth, height=0.4\linewidth]{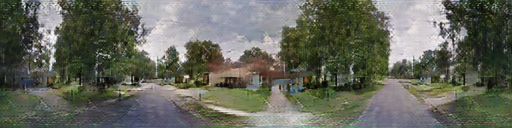} 
        \includegraphics[width=\linewidth, height=0.4\linewidth]{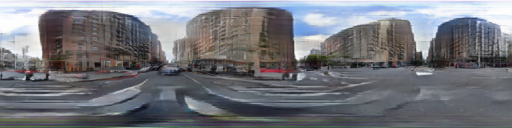} 
        \includegraphics[width=\linewidth, height=0.4\linewidth]{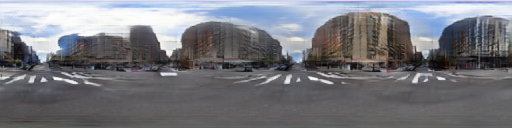} 
        \includegraphics[width=\linewidth, height=0.4\linewidth]{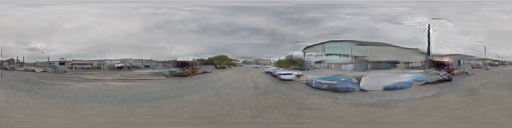}  
      \caption{\small Sat2Den}
    \end{subfigure}
    \begin{subfigure}{0.22\linewidth}
        \includegraphics[width=\linewidth, height=0.4\linewidth]{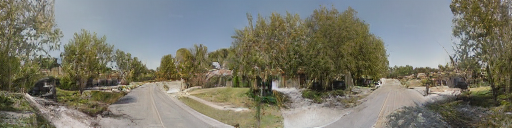} 
        \includegraphics[width=\linewidth, height=0.4\linewidth]{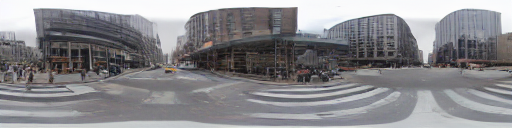} 
        \includegraphics[width=\linewidth, height=0.4\linewidth]{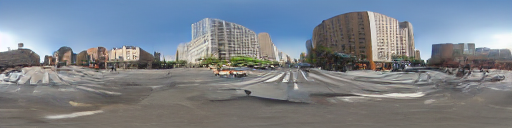} 
        \includegraphics[width=\linewidth, height=0.4\linewidth]{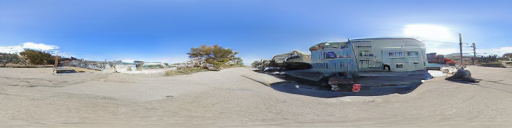}
      \caption{\small ControlS2S}
    \end{subfigure}
    \begin{subfigure}{0.22\linewidth}
        \includegraphics[width=\linewidth, height=0.4\linewidth]{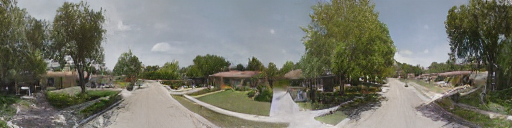}
        \includegraphics[width=\linewidth, height=0.4\linewidth]{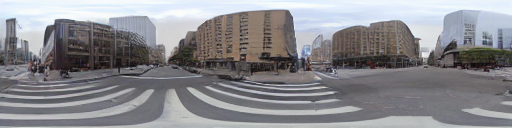} 
        \includegraphics[width=\linewidth, height=0.4\linewidth]{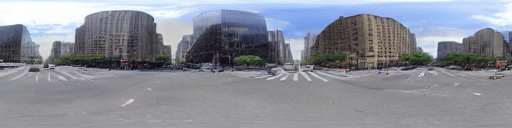} 
        \includegraphics[width=\linewidth, height=0.4\linewidth]{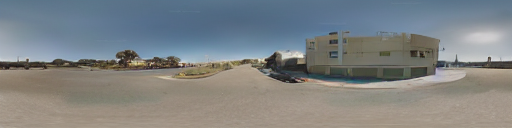}
      \caption{\small Ours}
    \end{subfigure}
    \begin{subfigure}{0.22\linewidth}
        \includegraphics[width=\linewidth, height=0.4\linewidth]{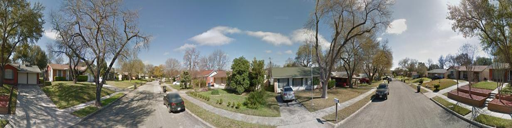} 
        \includegraphics[width=\linewidth, height=0.4\linewidth]{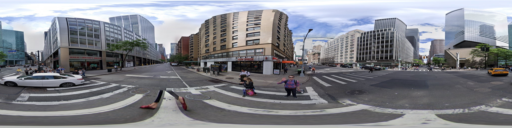} 
        \includegraphics[width=\linewidth, height=0.4\linewidth]{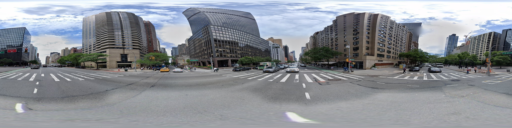} 
        \includegraphics[width=\linewidth, height=0.4\linewidth]{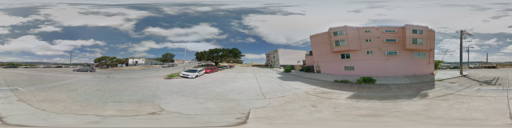}
      \caption{\small GT}
    \end{subfigure}
    \caption{\small Qualitative comparison with previous works on satellite to single ground image generation, our model can effectively capture roadways, ground markings, and architectural details.\label{fig:sigle_img_comp_sota}}
    \vspace{-1em}
  \label{fig:CVUSA_com}
  \end{figure*}
  \begin{table}[t]
  \centering
  \small
  \setlength{\abovecaptionskip}{0pt}
  \setlength{\belowcaptionskip}{0pt}
  \caption{\small Quantitative comparison with previous works on satellite to single ground image generation.\label{tab:sigle_img_comp_sota}}
  \setlength{\tabcolsep}{1.1mm}{
  \begin{tabular}{c|c|ccc|cc|ccc|c|c}
\midrule
\multirow{2}{*}{}                                         & \multirow{2}{*}{Method} & \multicolumn{3}{c|}{Perceptual Level}                                & \multicolumn{2}{c|}{Semantic Level}   & \multicolumn{3}{c|}{Pixel Level}                  & \multirow{2}{*}{$\downarrow$Depth} & \multirow{2}{*}{$\downarrow$LRCE} \\
                                                          &                         & $\downarrow$$P_{squeeze}$ & $\downarrow$$P_{alex}$ & $\downarrow$FID & $\downarrow$DINO & $\downarrow$SegAny & $\uparrow$SSIM  & $\uparrow$PSNR & $\uparrow$SD   &                                    &                       \\ \midrule
\multirow{6}{*}{\rotatebox{90}{CVUSA}} & Pix2Pix                 & 0.3468                    & 0.5084                 & 44.51           & 5.2415           & 0.3847             & 0.3190          & 13.20          & 12.08          & 21.85                              & 18.81                 \\
                                                          & S2S                     & 0.3218                    & 0.4830                 & 29.49           & 5.1117           & 0.3852             & 0.3508          & 13.40          & 12.30          & 21.05                              & 19.10                 \\
                                                          & Sat2Density             & 0.3217                    & 0.4634                 & 47.85           & 4.9445           & 0.3763             & 0.3307          & 13.46          & 12.27          & 19.83                              & 16.17                 \\
                                                          & CrossDiff               & -                         & -                      & 23.67           & -                & -                  & 0.3710          & 12.00          & -              & -                                  & -                     \\
                                                          & ControlS2S              & 0.3192                    & 0.4323                 & 21.30           & \textbf{4.807}   & 0.3612             & 0.3753          & 13.67          & 12.33          & 19.58                              & 14.62                 \\
                                                          & Ours                    & \textbf{0.3146}           & \textbf{0.4255}        & \textbf{17.00}  & \textbf{4.807}   & \textbf{0.3602}    & \textbf{0.3812} & \textbf{13.88} & \textbf{12.42} & \textbf{19.36}                     & \textbf{14.53}        \\ \midrule
\multirow{6}{*}{\rotatebox{90}{VIGOR}} & Pix2Pix                 & 0.3346                    & 0.4513                 & 67.96           & 4.717            & 0.3833             & 0.3714          & 13.33          & 12.93          & 8.647                              & 6.569                 \\
                                                          & S2S                     & 0.3694                    & 0.4941                 & 121.1           & 5.032            & 0.4037             & 0.3273          & 12.16          & 12.31          & 10.87                              & 9.790                 \\
                                                          & Sat2Density             & 0.2828                    & 0.3898                 & 54.49           & 4.408            & 0.3627             & 0.3956          & \textbf{14.14} & 12.38          & 8.054                              & 5.805                 \\
                                                          & ControlNet              & 0.3395                    & 0.4594                 & 23.68           & 4.950            & 0.3916             & 0.3397          & 12.02          & 12.59          & 10.02                              & 7.499                 \\
                                                          & ControlS2S              & 0.2729                    & 0.3770                 & 28.01           & 4.335            & 0.3529             & 0.4228          & 13.80          & 13.07          & 7.095                              & 5.176                 \\
                                                          & Ours                    & \textbf{0.2598}           & \textbf{0.3469}        & \textbf{21.36}  & \textbf{4.287}   & \textbf{0.3471}    & \textbf{0.4385} & 14.08          & \textbf{13.11} & \textbf{6.727}                     & \textbf{5.081}        \\ \midrule
\end{tabular}}
    \vspace{-2em}
  \end{table}
\subsection{Model Analysis}
Our method consists of two key components: (1) a ray-guided cross-view feature conditioning mechanism that ensures geometric consistency between the satellite image and the generated ground views, and (2) an epipolar-constrained attention module that enforces multi-view consistency across frames in the generated ground-view sequences.

To validate the effectiveness of the proposed \textbf{Ray-Guided Cross-view Feature Conditioning Mechanism}, we conduct single-image satellite-to-ground generation experiments on CVUSA and VIGOR, removing the influence of sequential modeling and allowing direct comparison with state-of-the-art methods. 
We compare our method with Pix2Pix~(\cite{pix2pix2017}), S2S~(\cite{shi2022geometry}), Sat2Density~(\cite{qian2023sat2density}), and ControlS2S~(\cite{ze2025controllable}), while CrossDiff~\cite{li2024crossviewdiff} results are cited from the original paper.
Note that pixel-wise metrics may not fully capture the quality of synthesized images in this task, as they are sensitive to factors such as lighting and sky appearance that are not explicitly modeled in satellite imagery.
Quantitative and qualitative comparisons on both datasets (Table~\ref{tab:sigle_img_comp_sota}, Figure~\ref{fig:sigle_img_comp_sota}) demonstrate that our method achieves superior generation quality and stronger left–right boundary consistency. This improvement arises from the proposed Ray-Based Pixel Attention module, which samples features along the viewing ray and explicitly incorporates geometric cues, enabling faithful alignment with the satellite-view representation.

Next, we verify the necessity of the proposed \textbf{Epipolar-Constrained Temporal Attention}. As shown in Figure~\ref{fig:ablation_vis}, incorporating this module significantly improves sequence consistency. Furthermore, Table~\ref{tab:temporal_consistency} and Figure~\ref{fig:GPU_memory} demonstrate that replacing Epipolar-Constrained Attention with full cross-attention substantially increases computational cost, whereas our approach achieves both lower cost and stronger sequence consistency. The advantage stems from introducing geometric priors via epipolar geometry, which filters out irrelevant matches, suppresses
noise propagation, and eliminates a large number of non-corresponding points before attention computation. Additionally, our sparse interframe attention, in which each frame attends only to its immediate neighbors, allows the model to scale efficiently to longer sequences without compromising performance.


\textbf{Application to Downstream Cross-View Localization Task.}
SatDream360 can be leveraged to generate synthetic ground-view data from satellite imagery, enabling enhanced training for downstream tasks. We evaluate this benefit in the context of cross-view localization using the state-of-the-art G2SWeakly~\cite{shi2024weakly} model as a baseline. To ensure fair comparison, we follow the same training configuration as the baseline: 10 epochs with identical batch sizes. The only modification is the inclusion of SatDream360-generated data for training augmentation. As shown in Table~\ref{tab:cross_localization}, the augmented model achieves superior performance, demonstrating that the high-fidelity, geometrically consistent samples produced by SatDream360 provide meaningful improvements for cross-view localization tasks.

\begin{table}[t]
    \centering
    \begin{minipage}{0.6\textwidth}
      \centering
      \includegraphics[width=\textwidth]{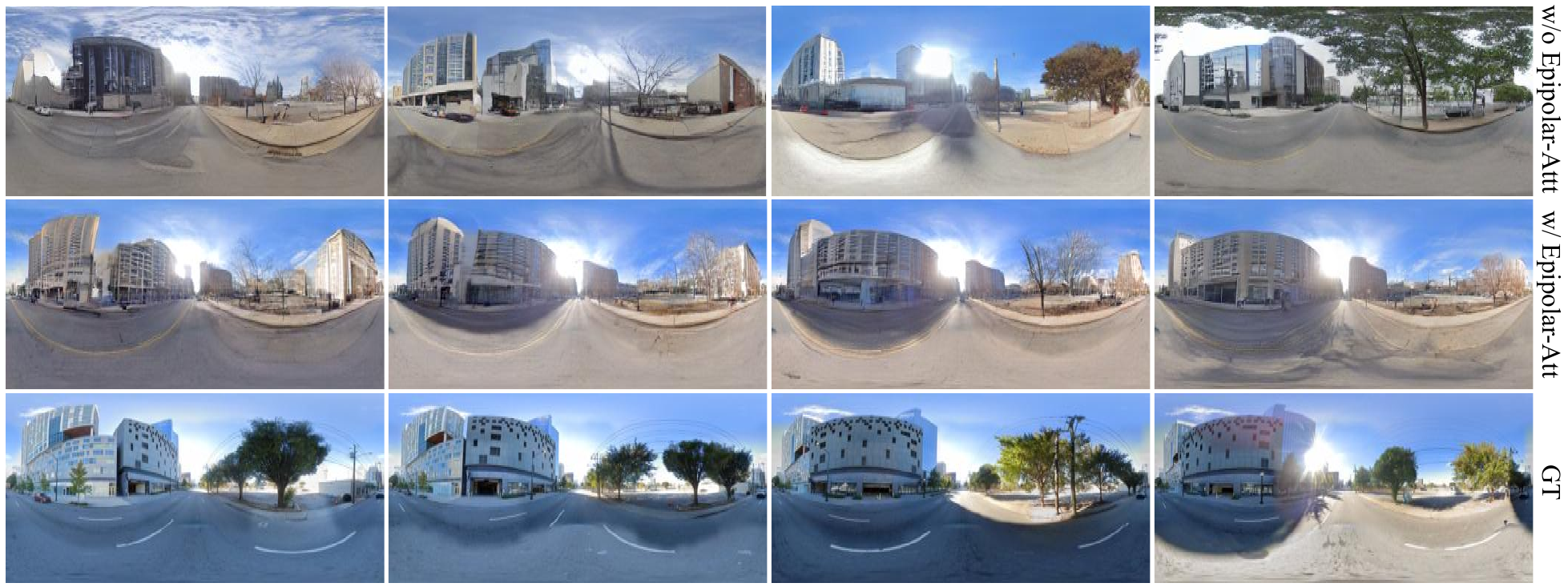} 
      \captionof{figure}{\small Qualitative comparison without (top) and with (middle) the proposed epipolar-constrained attention. \label{fig:ablation_vis}}
      \end{minipage}
      \hfill 
    \begin{minipage}{0.38\textwidth}
        \setlength{\abovecaptionskip}{0pt}
        \setlength{\belowcaptionskip}{0pt}
        \centering
        \caption{\small 
        Application to the downstream cross-view localization task. 
        Experimental evaluation on the VIGOR dataset reveals the average localization error before and after synthetic data training.
        \label{tab:cross_localization}}
        \setlength{\tabcolsep}{0.4mm}{
        \begin{tabular}{c|cc}
          \midrule
                             & $\downarrow$Aligned       & $\downarrow$Unaligned     \\ \midrule
          w/o synth data & 5.22          & 5.33          \\
          w/ Ours            & \textbf{4.99} & \textbf{5.11} \\ \midrule
          \end{tabular}}
    \end{minipage}%
    \vspace{-2em}
  \end{table}
\begin{table}[t]
    \centering
    \begin{minipage}{0.6\textwidth}
        \centering
        \setlength{\abovecaptionskip}{0pt}
        \setlength{\belowcaptionskip}{0pt}
        \vspace{1em}
        \caption{\small 
        Comparison of Full Cross-Attention and Epipolar-Constrained Temporal Attention for realism and temporal consistency.
        \label{tab:temporal_consistency}}
        \setlength{\tabcolsep}{0.5mm}{
        \begin{tabular}{c|ccc|cc}
            \midrule
                              & $\downarrow$FID           & $\downarrow$DINO          & $\downarrow$Depth         & $\downarrow$FVD           & $\downarrow$CLIPSIM        \\ \midrule
            w/ Full Cross-Att & 42.60          & 4.253          & 6.231          & 2.150          & 7.516          \\
            w/ Epipolar-Att  & \textbf{27.41} & \textbf{4.156} & \textbf{5.623} & \textbf{2.101} & \textbf{6.820} \\ \midrule
            \end{tabular}}
    \end{minipage}%
    \hfill 
    \begin{minipage}{0.38\textwidth}
        \vspace{0.3cm}
        \centering
        \includegraphics[width=\textwidth]{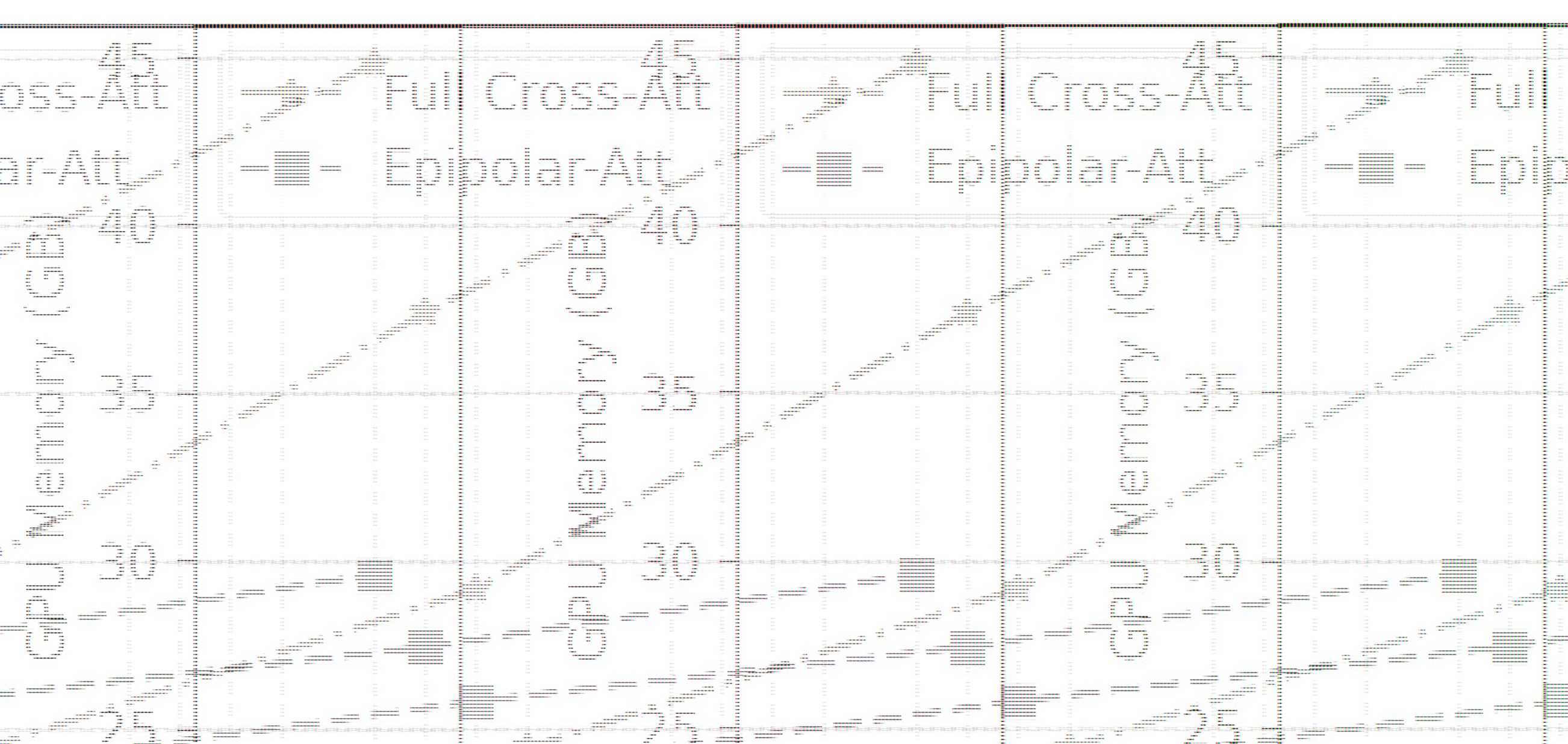} 
        \vspace{-2em}
        \captionof{figure}{\small Memory comparison when generating different frame numbers in a video.\label{fig:GPU_memory}}
    \end{minipage}
    \vspace{-2em}
  \end{table}
\section{Conclusion}
We propose a novel framework for satellite-to-ground multiview generation, addressing the challenging task of synthesizing continuous ground-level panoramas from a single top-down satellite image. Our approach tackles both spatial and multiview consistency through two key modules: (1) a Ray-Guided Cross-View Feature Conditioning mechanism for accurately constructing satellite-and-ground-view correspondences, and (2) a Multi-scale Epipolar-Constrained Attention module that ensures multiview consistency with significantly reduced computational cost compared to standard cross-attention. To facilitate evaluation, we introduce VIGOR++, a large-scale benchmark dataset of aligned panoramic sequences and satellite views. Extensive experiments across multiple metrics and datasets demonstrate that our method outperforms state-of-the-art baselines in perceptual realism, semantic consistency, and multiview stability. We believe that this work provides a strong foundation for future research in cross-view generative modeling, with broad potential applications in 3D reconstruction, autonomous driving, and simulation environments.

\subsubsection*{Reproducibility Statement}
The implementation details of our model are provided in Section~\ref{sec:Method}, with training settings and evaluation protocols provided in Section~\ref{sec:Experiments} and Appendix~\ref{sec:Experiments_det}. Additional ablation studies are included in the Appendix~\ref{sec:ab_1},~\ref{sec:ab_2},~\ref{sec:ab_3} to clarify the effect of individual components. We promise to release both the dataset and the code to facilitate reproducibility.



\bibliography{Ref.bib}
\bibliographystyle{iclr2026_conference}

\newpage
\appendix
\section{Appendix}
\subsection{The Use of Large Language Models (LLMs)}
Large Language Models (LLMs) were employed solely as writing and editing assistants during manuscript preparation. In particular, we used an LLM to refine language, improve readability, and enhance clarity across various sections. Its contributions included tasks such as sentence rephrasing, grammar correction, and improving the overall coherence and flow of the text.

Importantly, the LLM played no role in the conception of research ideas, methodological design, or experimental execution. All core concepts, analyses, and results were entirely developed and validated by the authors. The LLM’s involvement was strictly limited to linguistic refinement and did not influence the scientific content or data interpretation. The authors take full responsibility for the entirety of the manuscript, including any portions polished with LLM assistance. We have ensured that all LLM-generated content adheres to ethical standards and does not contribute to plagiarism or scientific misconduct.

\subsection{Explanation of Projection Geometry \label{App:Projection}}
\begin{minipage}{0.56\textwidth}
We provide an explanation of Eq.~\ref{eq:cam_project}, describing the correspondence between pixel coordinates in the camera coordinate system and the angles of the rays found.
Any point $(u,v)$ in the pixel coordinate system corresponds to a camera ray with angles $(\psi,\theta)$, where $\psi$ is the yaw angle ranging from $[-\pi,\pi]$ and $\theta$ is the pitch angle ranging from $[-\pi/2,\pi/2]$. For example, as shown in Figure~\ref{fig_coor}, the pixel at coordinates $(u',v')$ corresponds to the following ray angles:
\end{minipage}
\hfill
\begin{minipage}{0.4\textwidth}
    \vspace{-4em}
    \centering
    \includegraphics[width=0.9\textwidth]{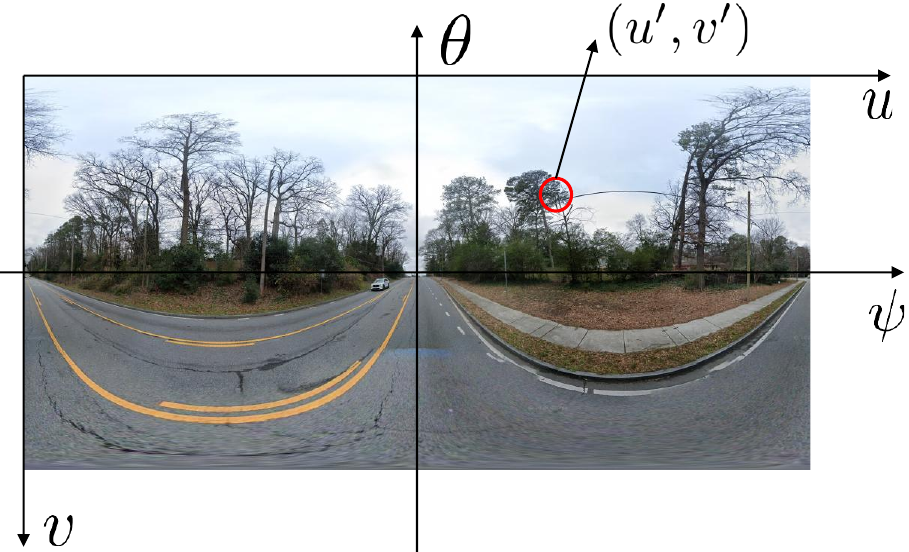}
    \captionof{figure}{\small
        Correspondence between image pixel coordinates and camera ray angles.
        \label{fig_coor}}
    \vspace{-1em}
\end{minipage}
\begin{equation}
    \psi_{u',v'} = (u' - \frac{W}{2})/W \times 2\pi , \quad
    \theta_{u',v'} = (\frac{H}{2} - v')/H \times \pi,
\end{equation}
where H and W are the height and width of the panoramic image.

\subsection{Proof of Eq.~\ref{eq:epipolar} \label{App:epipolar_proof}}
\textbf{Prepare.} For vectors $a = \left[a_1 \; a_2 \; a_3 \right]^T$ and $b = \left[b_1 \; b_2 \; b_3 \right]^T$, the outer product of the two vectors is defined as:
\begin{equation}
a \otimes b 
= \begin{bmatrix} a_2b_3 - a_3b_2 \\ a_3b_1 - a_1b_3 \\ a_1b_2 - a_2b_1 \end{bmatrix}
= \begin{bmatrix} 0 & -a_3 & a_2 \\ a_3 & 0 & -a_1 \\ -a_2 & a_1 & 0 \end{bmatrix}b
= \hat{a}b \label{eq:prepare}.
\end{equation}
Here, we introduce the skew-symmetric matrix notation, $\hat{a}$, which corresponds to the vector $a$. Using this notation, the outer product $a \otimes b $ is expressed compactly as the matrix-vector product $\hat{a}b$.

\textbf{Proof.} Consider the ground images of frames $m$ and $n$, there is a spatial point $Q=(x,y,z)$ in the coordinate system of frame $m$ corresponds to pixel points  $g^m_{u,v}$ on frame $m$ and $g^n_{u',v'}$ on frame $n$ in panoramic images: These projections are given by:
\begin{equation}
    g^m_{u,v} = P(Q), \quad
    g^n_{u',v'} = P(R_{mn}Q + t_{mn}),
\end{equation}
where $P$ denotes the equirectangular camera projection transformation, $R_{mn}$ and $t_{mn}$ denote the relative rotation and translation between frames $m$ and $n$. Perform the inverse projection transformation to obtain:
\begin{equation}
    P^{-1}(g^m_{u,v}) = Q, \quad
    P^{-1}(g^n_{u',v'}) = R_{mn}Q + t_{mn}.
\end{equation}
Substitute $Q$ from the left equation into the right gives:
\begin{equation}
    P^{-1}(g^n_{u',v'}) = R_{mn}(P^{-1}(g^m_{u,v})) + t_{mn}.
\end{equation}
By simultaneously left multiplying with skew-symmetric matrix $\hat{t}{mn}$ (corresponding to $t_{mn}$), as introduced in Eq.~\ref{eq:prepare}:
\begin{equation}
    \hat{t}_{mn}(P^{-1}(g^n_{u',v'})) = \hat{t}_{mn}R_{mn}(P^{-1}(g^m_{u,v})).
\end{equation}
Next, we multiply both sides on the left by the transpose  $(p^{-1}(g^n_{u',v'}))^T$:
\begin{equation}
    (P^{-1}(g^n_{u',v'}))^T\hat{t}_{mn}(P^{-1}(g^n_{u',v'})) = (P^{-1}(g^n_{u',v'}))^T\hat{t}_{mn}R_{mn}(P^{-1}(g^m_{u,v})).
\end{equation}
Since the product$(P^{-1}(g^n_{u',v'}))^T\hat{t}_{mn}(P^{-1}(g^n_{u',v'})) = 0$ (because $\hat{t}_{mn}(P^{-1}(g^n_{u',v'}))$ is orthogonal to  $P^{-1}(g^n_{u',v'})$), the above simplifies to:
\begin{equation}
    (P^{-1}(g^n_{u',v'}))^T\hat{t}_{mn}R_{mn}(P^{-1}(g^m_{u,v})) = 0.
\end{equation}
This result implies that pixels $g^m_{u,v}$ on frame $m$ and $g^n_{u',v'}$ on frame $n$ that correspond to the same spatial point must satisfy this constraint relationship in Eq.~\ref{eq:epipolar}. Therefore, during temporal attention, the points $g^m_{u,v}$ on frame $m$ only need to focus on the set of points $\{g^n_{u',v'}\}$ on frame $n$ that satisfy the above constraint. This significantly reduces the computational complexity compared to focusing on all pixels in the image.

\subsection{Scaling to Large-Scale Scenes\label{App:triplane_updare}}
\begin{figure}[h]
    \centering  
    \setlength{\abovecaptionskip}{0pt}
    \setlength{\belowcaptionskip}{0pt}
    \includegraphics[width=1\textwidth]{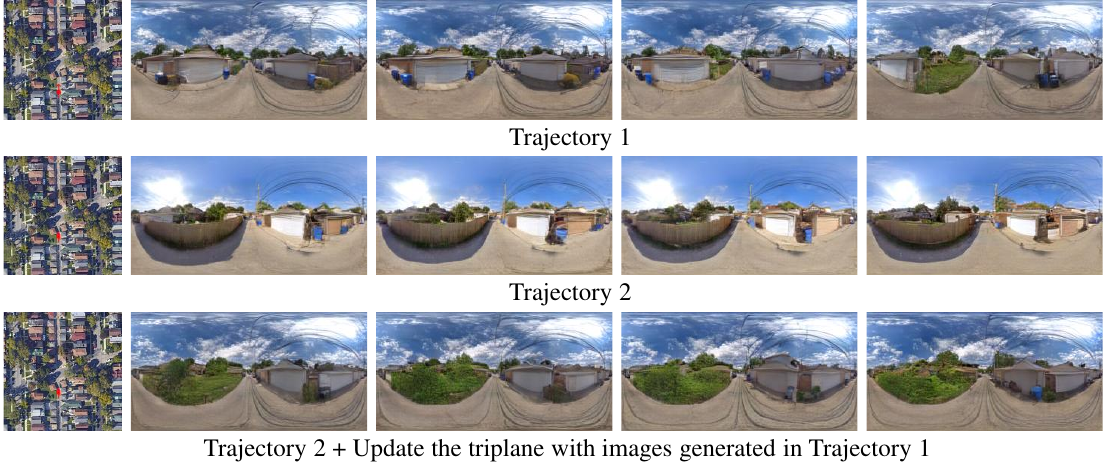}
    \caption{\small
        Using the same satellite image as a condition, different trajectories are input to generate corresponding images. From top to bottom: results for trajectory 1, results for trajectory 2, and results for trajectory 2 after updating the triplane with images generated from trajectory 1. Updating the triplane ensures that newly generated results are related to prior sequences.
        \label{fig:triplane}}
    \vspace{-1.3em}
\end{figure}

As described in Section~\ref{sec:Method}, the proposed model leverages Ray-Guided Cross-View Feature Conditioning and Epipolar-Constrained Attention to generate ground-view images aligned with the satellite inputs and consistent across views. However, scaling the approach to larger scenes remains challenging. Due to the inherent stochasticity of diffusion models, different batches conditioned on the same satellite image can produce noticeably different results, as illustrated in the first two rows of Figure~\ref{fig:triplane}. This highlights the need to establish connections across images generated from multiple batches to enhance the scalability of scene generation.

To incorporate information from previously generated images into the upcoming sequence, we update the triplane representation using Image Cross-Attention (ICA~\cite{li2024bevformer}), which enriches each point on the XY, XZ, and YZ planes by referencing corresponding pixels from previously generated images. For example, points on the XY plane are sampled along the $Z$-axis, projected into camera space, and aligned with image coordinates, enabling feature transfer through ICA: 
\begin{equation}
    F_{xy}^{\text{top}} = \mathrm{ICA}\left(F_{xy}^{\text{top}}, \mathrm{Ref}_{xy}^{2D}\right), \quad 
    \mathrm{Ref}_{xy}^{2D} = \{F^j_{u_i,v_i}\}.
\end{equation}

Here, $(u_i , v_i) = P(R_j(x, y, z_i)+ t_j)$ projects sampled 3D points into the $j$-th image plane via rotation $R_j$, translation $t_j$, and projection $P$, where $\{(x, y, z_i)\}$ represents the set of sampled points along the $Z$-axis. By integrating ICA with CVHA, the triplane accumulates information from prior results, yielding coherent and scalable scene generation across multiple batches (Figure~\ref{fig:triplane}, bottom row). By combining ICA and CVHA mechanisms, the triplane's features are enriched through the aggregation of information from previously generated images. As shown at the bottom of Figure~\ref{fig:triplane}, this approach enables the generation of extended image sequences across multiple batches, ensuring coherence and continuity in the outputs.


\subsection{Additional Experiment Details\label{sec:Experiments_det}}
Our model is fine-tuned based on the pre-trained Stable Diffusion 1.5 model~(\cite{rombach2022high}), with the training process divided into three stages. 

Initially, we train on single-image generation tasks with a batch size of 32 for 300 epochs, focusing on adapting the parameters of Ray-Guided Cross-View Feature Conditioning to generate ground-level images consistent with satellite geometry. 

Next, we incorporate the Epipolar-Constrained Temporal Attention module to enable continuous ground scene generation This phase involves a total of 300 epochs. The first 5 epochs focus on pre-heating the Temporal Attention module with a batch size of 8 and 3 images per sequence. This is followed by fine-tuning the entire model for 200 epochs with the same batch size and sequence length, and then further fine-tuning for long sequence generation with a batch size of 4 and 5 images per sequence until reaching 300 epochs.

As the Autoencoder was originally trained only on single-image tasks, it caused flickering artifacts when handling a temporally coherent sequence of images. Same to~(\cite{blattmann2023align}), we maintain the original parameters and introduce temporal modules (3D convolution) in the decoder. The Autoencoder is trained on VIGOR++ with a batch size of 4 and 3 images per sequence for 40 epochs, focusing solely on training the added temporal module.

Throughout the training process, the learning rate is set to 7.0e-05, the optimizer used is AdamW, and all experiments are conducted on four NVIDIA L40 GPUs. This comprehensive training pipeline enables the model to generate geometrically accurate ground-level images and seamless, temporally coherent sequences.

\subsection{Ablation Study on Scene Representations \label{sec:ab_1}}
The comparison between triplane and BEV scene representations is presented in Table~\ref{tab:ab_triplan}, where the BEV corresponds to using only the XY-plane from the triplane.
\begin{table}[h]
  \centering
  \small
  \setlength{\abovecaptionskip}{0pt}
  \setlength{\belowcaptionskip}{0pt}
  \caption{\small Quantitative comparison of Triplane and BEV(XY-plane)  representations on the VIGOR dataset.\label{tab:ab_triplan}}
  \setlength{\tabcolsep}{1.4mm}{
 \begin{tabular}{c|ccccccc}
\midrule
            & $\downarrow$$P_{alex}$      & $\downarrow$DINO           & $\downarrow$SegAny          & $\uparrow$SSIM            & $\uparrow$PSNR           & $\uparrow$SD             & $\downarrow$Depth           \\ \midrule
BEV & 0.3803          & 4.408          & 0.3549          & 0.4134          & 13.64          & 12.94          & 7.061          \\
triplane  & \textbf{0.3469} & \textbf{4.287} & \textbf{0.3471} & \textbf{0.4385} & \textbf{14.08} & \textbf{13.11} & \textbf{6.727} \\ \midrule
\end{tabular}}
\vspace{-1em}
\end{table}

These results show that the triplane representation consistently outperforms the BEV(XY-plane) representation across all metrics. The improvement comes from the triplane’s ability to capture richer 3D structures. A pure XY-plane representation lacks vertical information, which is critical for rendering views under varying pitch angles. Moreover, our model samples features along camera rays. Relying solely on XY-plane features leads to incomplete spatial support, especially for oblique rays. The triplane effectively overcomes this limitation.

\subsection{Ablation Study on Ray-Based Pixel Attention \label{sec:ab_2}}
The standard method for applying triplane representations as diffusion conditions involves performing cross-attention between every 3D point feature encoded in the triplane and each image pixel. However, this approach lacks explicit geometric constraints, often results in geometric distortions, and incurs heavy memory consumption.

\begin{table}[h]
  \centering
  \small
  \setlength{\abovecaptionskip}{0pt}
  \setlength{\belowcaptionskip}{0pt}
  \caption{\small Quantitative comparison of Standard Cross-Attention and Ray-Based Pixel Attention on the CVUSA dataset. \label{tab:Ab_ray_att}}
  \setlength{\tabcolsep}{0.9mm}{
  \begin{tabular}{c|ccccccccc}
\midrule
                          & $\downarrow$$P_{alex}$      & $\downarrow$DINO           & $\downarrow$SegAny          & $\uparrow$SSIM            & $\uparrow$PSNR           & $\uparrow$SD             & $\downarrow$Depth          & $\downarrow$Time(s)        & $\downarrow$Memory(MB)     \\ \midrule
Standard Cross Att  & 0.5413          & 5.425          & 0.3911          & 0.3174          & 12.35          & 12.05          & 25.38          & 120.28         & 22506          \\
Ray-Based Pixel Att & \textbf{0.3469} & \textbf{4.287} & \textbf{0.3471} & \textbf{0.4385} & \textbf{14.08} & \textbf{13.11} & \textbf{6.727} & \textbf{39.64} & \textbf{18296} \\ \midrule
\end{tabular}}
  \end{table}

To address this, we propose Ray-Based Pixel Attention, which dynamically samples points along viewing rays by learnable offsets (Eq.~\ref{eq:Ray_att}). This approach ensures spatial coherence and improves geometric alignment across views while significantly reducing memory usage. Table~\ref{tab:Ab_ray_att} reports comparison results on the CVUSA dataset, demonstrating that Ray-Based Pixel Attention achieves higher generation quality with lower resource consumption. Both Time and Memory are measured with a batch size of 30 and 50 DDIM steps.

Furthermore, we conduct an ablation study on the Dynamic Refinement of Offsets in Ray-Based Pixel Attention, with results shown in Table~\ref{tab:Dy_off}:
\begin{table}[h]
  \centering
  \small
  \setlength{\abovecaptionskip}{0pt}
  \setlength{\belowcaptionskip}{0pt}
  \caption{\small Ablation study on the effect of dynamic refinement tested on the CVUSA dataset\label{tab:Dy_off}}
  \setlength{\tabcolsep}{1.4mm}{
 \begin{tabular}{c|ccccccc}
\midrule
            & $\downarrow$$P_{alex}$      & $\downarrow$DINO           & $\downarrow$SegAny          & $\uparrow$SSIM            & $\uparrow$PSNR           & $\uparrow$SD             & $\downarrow$Depth          \\ \midrule
w/o Dynamic & 0.4647          & 4.935          & 0.3639          & 0.3736          & 13.45          & 12.38          & 19.69          \\
w/ Dynamic  & \textbf{0.4255} & \textbf{4.807} & \textbf{0.3602} & \textbf{0.3812} & \textbf{13.88} & \textbf{12.42} & \textbf{19.36} \\ \midrule
\end{tabular}}
    \vspace{-1em}
  \end{table}
  
The goal of Ray-Based Pixel Attention is to aggregate meaningful features for each ground-level pixel by sampling along its corresponding 3D ray projected into the triplane. However, Early in the diffusion process, the latent features are dominated by noise, making accurate 3D correspondence difficult. To address this, we begin with uniform sampling along each ray. But uniform sampling often leads to sparse or suboptimal feature aggregation, leading to degraded performance, as shown in the first row of Table~\ref{tab:Dy_off}.

To mitigate this, we propose a mechanism to dynamically refine both the sampling offsets and their corresponding weights during diffusion inference. As denoising progresses, latent features gradually capture meaningful scene structure. The model uses these evolving features to predict offsets and weights for each sampling point along the ray, enabling more accurate and view-consistent feature aggregation. This leads to improved performance, as shown in the second row of Table~\ref {tab:Dy_off}.

\subsection{Ablation Study on sparse interframe attention \label{sec:ab_3}}
We adopt a sparse interframe strategy after careful comparison with the dense interframe strategy. Specifically, the sparse strategy queries only the two preceding frames for each target frame, whereas the dense strategy attends to all frames within the sequence. As shown in Table~\ref{tab:sparse_value}, both strategies achieve similar per-frame quality, but the sparse strategy yields better multiview consistency. This is because distant frames often contribute less meaningful information and may introduce noise, while nearby frames provide more relevant context.
\begin{table}[h]
  \centering
  \small
  \setlength{\abovecaptionskip}{0pt}
  \setlength{\belowcaptionskip}{0pt}
  \caption{\small Ablation study on the effect of sparse interframe strategy tested on the VIGOR++ dataset.\label{tab:sparse_value}}
  \setlength{\tabcolsep}{1.4mm}{
\begin{tabular}{c|cccccc}
\midrule
       & $\downarrow$FVD            & $\downarrow$CLIPSIM        & $\downarrow$$P_{alex}$            & $\downarrow$DINO           & $\downarrow$Depth          & $\uparrow$PSNR           \\ \midrule
Dense  & 2.253          & 7.071          & 0.4085          & \textbf{4.153} & 5.791          & \textbf{12.91} \\
Sparse & \textbf{2.101} & \textbf{6.820} & \textbf{0.3955} & 4.156          & \textbf{5.623} & 12.75          \\ \midrule
\end{tabular}}
\vspace{-1em}
\end{table}

We further compare runtime and memory usage as the number of frames increases (with batch size set to 1 and 50 DDIM steps). Table~\ref{tab:sparse_time} shows that the sparse strategy is significantly more efficient, particularly for longer sequences.

\begin{table}[h]
  \centering
  \small
  \setlength{\abovecaptionskip}{0pt}
  \setlength{\belowcaptionskip}{0pt}
  \caption{\small Ablation study on the resource consumption of the sparse interframe strategy.\label{tab:sparse_time}}
  \setlength{\tabcolsep}{1.4mm}{
\begin{tabular}{c|c|ccc}
\midrule
Strategy & Metric & 10 Frames & 20 Frames & 30 Frames \\ \midrule
Dense             & Time(s)         & 20.33              & 59.12              & 120.75             \\
                  & Memory(MB)      & 26174              & 31106              & 40520              \\ \midrule
Sparse            & Time(s)         & \textbf{14.29}     & \textbf{22.50}     & \textbf{32.71}     \\
                  & Memory(MB)      & \textbf{25824}     & \textbf{30242}     & \textbf{35142}     \\ \midrule
\end{tabular}}
\vspace{-1em}
\end{table}

In summary, the sparse inter-frame strategy achieves better multiview consistency while reducing both computation time and memory usage, making it the preferred choice for our generation framework.

\subsection{Limitations and Discussions}
While SatDreamer360 can generate continuous ground scene images from a single satellite image and a given ground camera trajectory, it still faces several limitations and corresponding areas for improvement.

Although VIGOR++ covers diverse regions, it is still constrained by Google Maps coverage and existing road networks, which may limit generalization to off-road or unstructured environments. Future work will incorporate additional data sources such as drone imagery, vehicle-mounted cameras, or crowd-sourced panoramic data to expand coverage and enhance the model’s generalization ability.

Additionally, when two poses in a trajectory are very close ($<$8m), the model may generate nearly identical images. This issue arises from the sparse sampling in the training data (15–20m) and limited triplane resolution. The issue disappears when the inter-frame distance exceeds 8m and can be mitigated by increasing triplane resolution, adding more densely sampled trajectories during training, or applying frame interpolation~(\cite{xing2024dynamicrafter,wan2025}) to the generated sequence. 

Finally, the inference process remains slow due to the multiple sampling steps required by the diffusion model. This limitation can be alleviated by exploring more efficient diffusion architectures and distillation-based acceleration techniques.


\subsection{Broader Social Impacts}
Our proposed system, SatDreamer360, can generate continuous ground scene images from a single satellite image and a given ground camera trajectory, making it a valuable tool for applications such as 3D reconstruction, simulation, and autonomous navigation. However, while it can produce visually plausible ground scenes, it still struggles to capture all real-world details, and caution should be exercised when deploying it in safety-critical scenarios.

Moreover, like many generative models, SatDreamer360 could be misused to synthesize misleading or fake visual content. To mitigate such risks, we recommend using it only in controlled research or industrial settings, incorporating usage licenses and watermarking techniques to trace generated content, and clearly disclosing when images are synthetic. These safeguards can help prevent misuse and ensure that the technology is applied responsibly.

\subsection{Examples of Satellite Images and Trajectories}
\begin{figure}[h]
    \centering
    \setlength{\abovecaptionskip}{0pt}
    \setlength{\belowcaptionskip}{0pt}
    \includegraphics[width=0.7\textwidth]{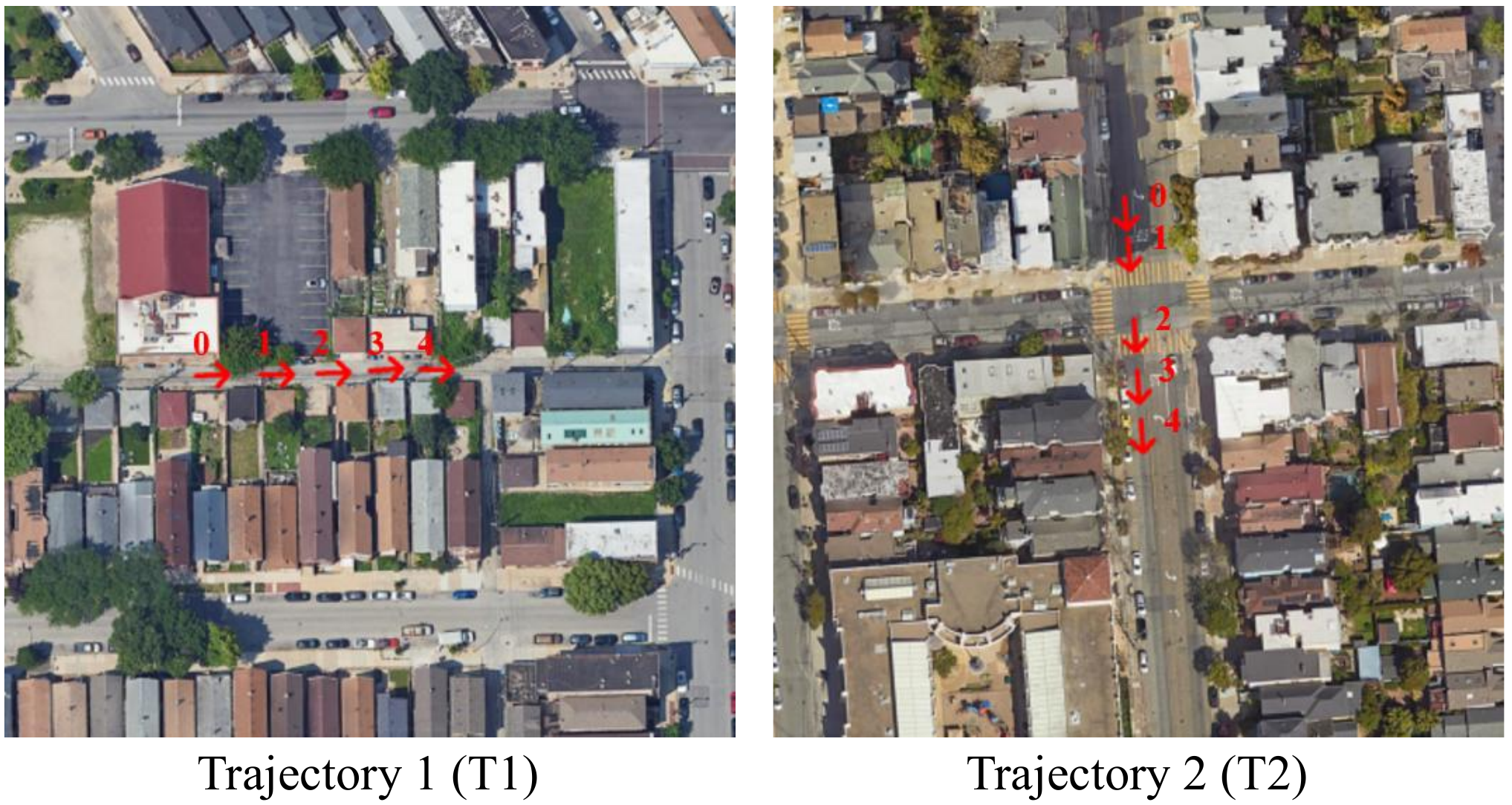}
    \caption{\small
        Satellite images and corresponding trajectory points \label{fig:sat_traject}}
\end{figure}

\subsection{More visualization results}
\begin{figure}[th]
\centering
\setlength{\abovecaptionskip}{0pt}
\setlength{\belowcaptionskip}{0pt}
\includegraphics[width=1\textwidth]{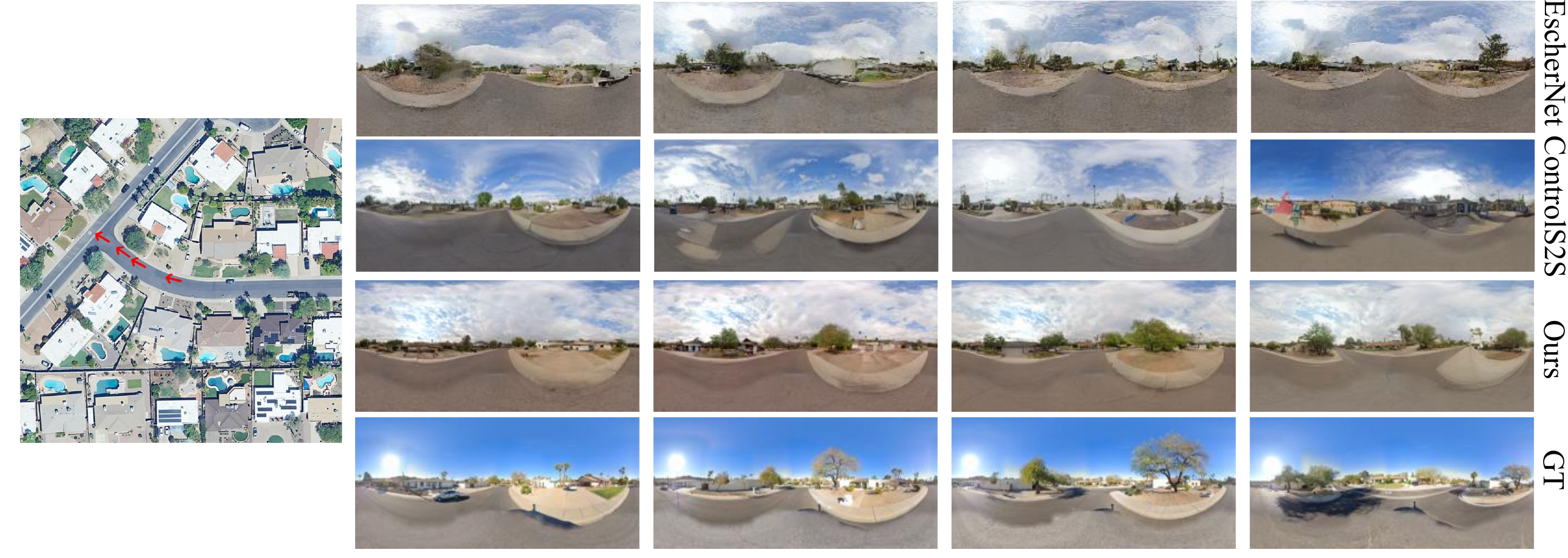}
\includegraphics[width=1\textwidth]{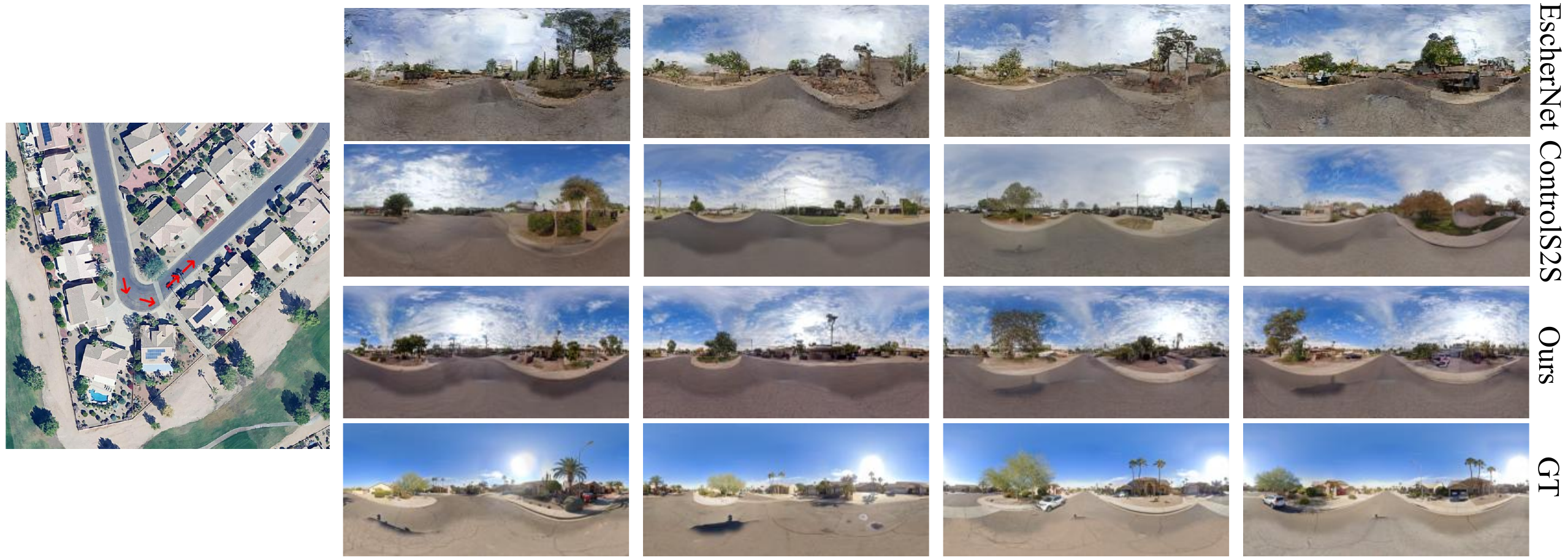}
\includegraphics[width=1\textwidth]{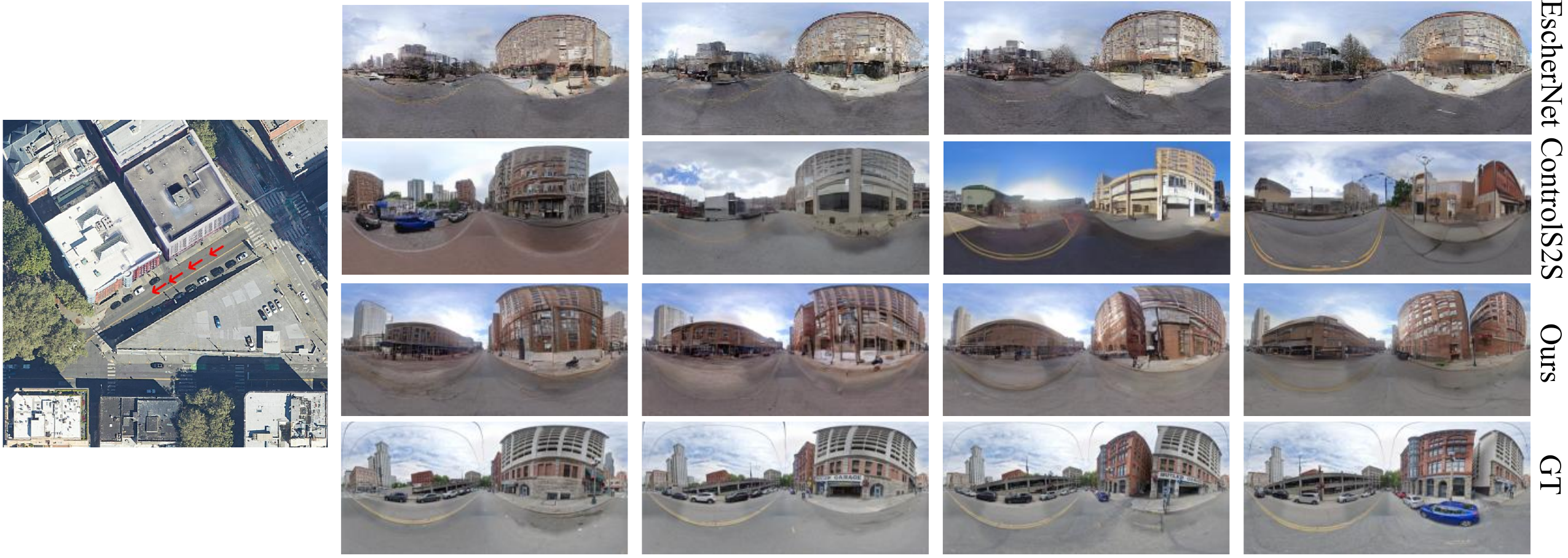}
\caption{\small
Qualitative comparison of ground-level image sequences along trajectories. The left shows the satellite image and the corresponding trajectory, while the ground-level images progress along the trajectory from left to right. \label{app_comp}}
\end{figure}

In Figure~\ref{app_comp}, we provide additional visualization results. Our method accurately follows the geometric layout of the satellite imagery while maintaining strong multiview consistency. It achieves robust performance both in densely built urban areas and rural regions. Even in challenging turning scenarios (middle of the figure), the generated results exhibit good continuity.

\end{document}